\documentclass[sn-mathphys-num]{sn-jnl}% Math and Physical Sciences Numbered Reference Style /
\RequirePackage{amsthm}
\usepackage{lscape}
\usepackage{changepage}
\usepackage{tabularx} % 支持自动调整宽度的表格
\usepackage{lipsum} % 示例文本
\usepackage{adjustbox} % 调整表格宽度的宏包
\usepackage{balance}
\usepackage{graphicx}
\usepackage{hyperref}
\usepackage{lscape}
\usepackage{longtable}
\usepackage{graphicx}%
\usepackage{multirow}%
\usepackage{amsmath,amssymb,amsfonts}%
\usepackage{amsthm}%
\usepackage{mathrsfs}%
\usepackage[title]{appendix}%
\usepackage{xcolor}%
\usepackage{textcomp}%
\usepackage{manyfoot}%
\usepackage{booktabs}%
\usepackage{algorithm}%
\usepackage{algorithmicx}%
\usepackage{algpseudocode}%
\usepackage{listings}%
\usepackage{tabularx}%
\usepackage[utf8]{inputenc}%
\UseRawInputEncoding

\theoremstyle{thmstyleone}%
\theoremstyle{thmstyletwo}%

\theoremstyle{thmstylethree}%

\begin{document}

\title[Article Title]{Deep Learning in Motion Deblurring: Current Status, Benchmarks and Future Prospects}

%%=============================================================%%
%% GivenName	-> \fnm{Joergen W.}
%% Particle	-> \spfx{van der} -> surname prefix
%% FamilyName	-> \sur{Ploeg}
%% Suffix	-> \sfx{IV}
%% \author*[1,2]{\fnm{Joergen W.} \spfx{van der} \sur{Ploeg} 
%%  \sfx{IV}}\email{iauthor@gmail.com}
%%=============================================================%%

\author[1]{\fnm{Yawen} \sur{Xiang}} % \email{ywxiang@bjtu.edu.cn}

\author[2]{\fnm{Heng} \sur{Zhou}} % \email{hengzhou@stu.xidian.edu.cn}
%\equalcont{These authors contributed equally to this work.}

\author[3,1]{\fnm{Chengyang} \sur{Li}} % \email{chengyang$\_$li@stu.pku.edu.cn}
%\equalcont{These authors contributed equally to this work.}
\author[1]{\fnm{Fangwei} \sur{Sun}} % \email{sun19961208@email.swu.edu.cn}
%\equalcont{These authors contributed equally to this work.}
\author*[1]{\fnm{Zhongbo} \sur{Li}}\email{zbli2021@outlook.com}
\author*[1]{\fnm{Yongqiang} \sur{Xie}}\email{yqxie2021@outlook.com}

\affil[1]{\orgdiv{Institute of Systems Engineering}, \orgname{Academy of Military Science}, \city{Beijing}}
\affil[2]{\orgdiv{School of Artificial Intelligence and Computer Science}, \orgname{Jiangnan University}, \city{Wuxi},  \country{China}}
\affil[3]{\orgdiv{School of Electronics Engineering and Computer Science}, \orgname{Peking University}, \city{Beijing}, \country{China}}

%\affil[2]{\orgdiv{Department}, \orgname{Organization}, \orgaddress{\street{Street}, \city{City}, \postcode{10587}, \state{State}, \country{Country}}}

%%==================================%%
%% Sample for unstructured abstract %%
%%==================================%%

\abstract{	Motion deblurring is one of the fundamental problems of computer vision and has received continuous attention.
	The variability in blur, both within and across images, imposes limitations on non-blind deblurring techniques that rely on estimating the blur kernel. As a response, blind motion deblurring has emerged, aiming to restore clear and detailed images without prior knowledge of the blur type, fueled by the advancements in deep learning methods.
	Despite strides in this field, a comprehensive synthesis of recent progress in deep learning-based blind motion deblurring is notably absent. This paper fills that gap by providing an exhaustive overview of the role of deep learning in blind motion deblurring, encompassing datasets, evaluation metrics, and methods developed over the last six years.
	Specifically, we introduce the types of motion blur and outline the shortcomings of traditional non-blind deblurring algorithms, emphasizing the advantages of employing deep learning techniques for deblurring tasks.
	Following this, we categorize and summarize existing blind motion deblurring methods based on different backbone networks, including convolutional neural networks, generative adversarial networks, recurrent neural networks, and Transformer networks.
	Subsequently, we elaborate not only on the fundamental principles of these different categories but also provide a comprehensive summary and comparison of their advantages and limitations. Qualitative and quantitative experimental results conducted on four widely used datasets further compare the performance of state-of-the-art methods.
	Finally, an analysis of present challenges and future prospects is intended to foster advancements and drive innovation in the field of image deblurring research.
	This review aspires to serve as a reference for those immersed in this field of study. All collected models, benchmark datasets, source code links, and codes for evaluation have been made publicly available at \url{https://github.com/VisionVerse/Blind-Motion-Deblurring-Survey}.
	}


%%\pacs[JEL Classification]{D8, H51}

%%\pacs[MSC Classification]{35A01, 65L10, 65L12, 65L20, 65L70}

\maketitle

\section{Introduction}\label{sec1}

%Motion blur is a common artifact in images caused by the relative movement between the camera and the scene during the exposure~\cite{zhang2021exposure,zhuang2023multi}.This effect can cause object contours in the image to be blurred or stretched, thereby reducing the clarity and detail of the image and affecting various computer vision tasks such as autonomous driving~\cite{rengarajan2016image,zhou2022fsad}, object segmentation~\cite{rajagopalan2023improving,guo2019multiview},medical image processing\cite{dai2021deep,li2023dsmt} and scene analysis~\cite{sheng2021improving}. Understanding the origins of motion blur is critical to developing effective elimination strategies.

%{In daily life, it is difficult to keep the equipment stable when taking images, and motion blur can easily occur due to camera shake or movement of the foreground target. As mentioned before, it not only affects the visual effect of images, but also poses a challenge to a series of computer vision tasks. To address this challenge, we need to construct a model that can describe the process of motion blur, thus providing us with a framework for understanding the nature of motion blur.}
{Motion blur is a common artifact in images caused by the relative movement between the camera and the scene during the exposure~\cite{zhang2021exposure,zhuang2023multi}. This effect can cause object contours in the image to be blurred or stretched, thereby reducing the clarity and detail of the image. It not only affects the visual effect of images, but also poses a challenge to a series of computer vision tasks such as autonomous driving~\cite{rengarajan2016image,zhou2022fsad}, object segmentation~\cite{rajagopalan2023improving,guo2019multiview}, medical image processing\cite{dai2021deep,li2023dsmt} and scene analysis~\cite{sheng2021improving}. 
To address this challenge, we need to construct a model that can describe the process of motion blur, thus providing us with a framework for understanding the nature of motion blur. }

\begin{figure}[h]
	\centering 
	\includegraphics[width=12cm]{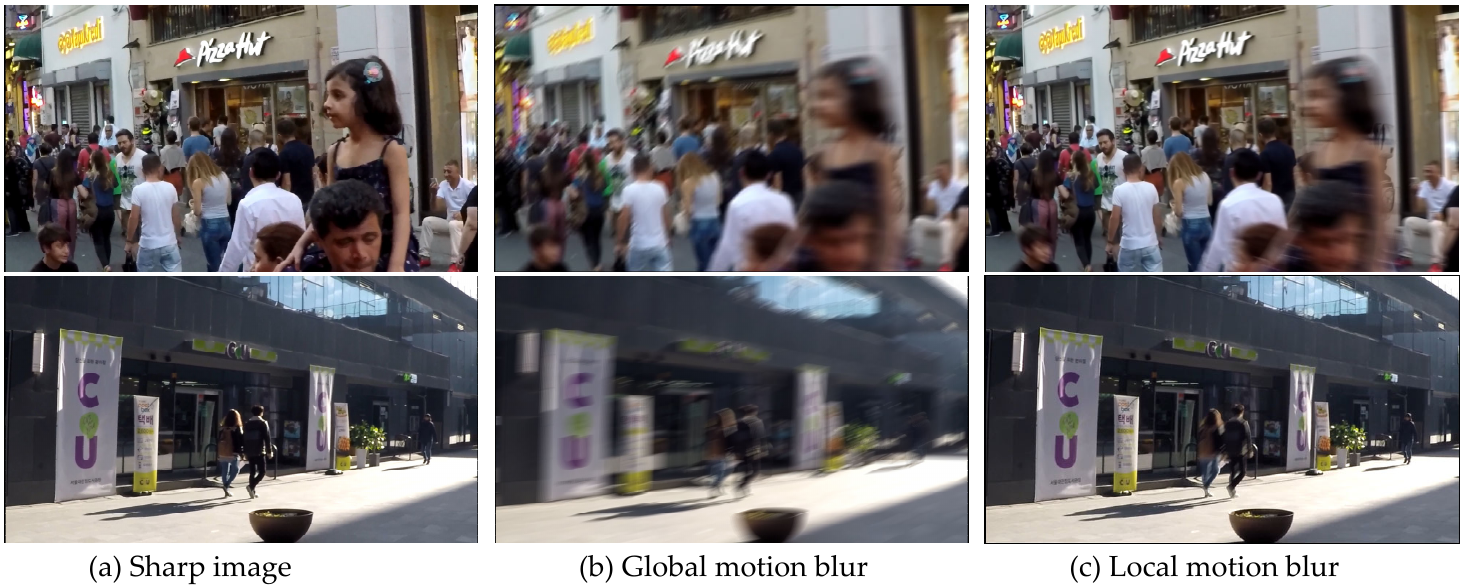}
	\caption{Sharp-blurry images.}
	\label{sharp_blur_image}
\end{figure}

The blurry process, as a fundamental degradation model, can be depicted as the convolution between a clear image and a blur kernel~\cite{cui2017high,bai2019single,richmond2022non}:
\begin{equation}
	\mathbf{B} = \mathbf{S} \otimes \mathbf{k} + \mathbf{n},
\end{equation}
where $ \mathbf{B} $ and $ \mathbf{S} $ are the observed blurred image and sharp image, respectively.
$ \mathbf{k} $ and $ \mathbf{n} $ represent the blur kernel and noise.
$ \otimes $ is the convolution operation.
{When blur kernel $ \mathbf{k} $ is known, the inverse process of this task is referred to as non-blind deblurring~\cite{arjomand2017deep,sun2014good,zhang2017learning,sheng2019depth,dong2018learning,dong2022dwdn,dong2021learning}.
However, $ \mathbf{k} $ is often unknown when dealing with blurred images, leading to tasks known as blind deblurring~\cite{zhang2024joint,pan2016blind,chen2019blind,hu2014deblurring,yan2017image,liu2019surface,santos2023deblur,pan2017learning}.
}
Motion blur can be divided into two types: global motion blur and local motion blur. As shown in \figurename~\ref{sharp_blur_image} (b), global motion blur is caused by the shaking of the camera~\cite{zhai2023comprehensive}, resulting in consistent blurring across the entire image. As shown in \figurename~\ref{sharp_blur_image} (c), local motion blur occurs due to relative motion between the target object and the static background~\cite{zhang2022deep}.

\begin{figure}[t]
	%	\begin{adjustwidth}{-\extralength}{0cm}
	\centering  % ????
		%		\includegraphics[width=14cm]{Definitions/Blind_deblur.pdf}
		%	\end{adjustwidth}
	\includegraphics[width=11.5cm]{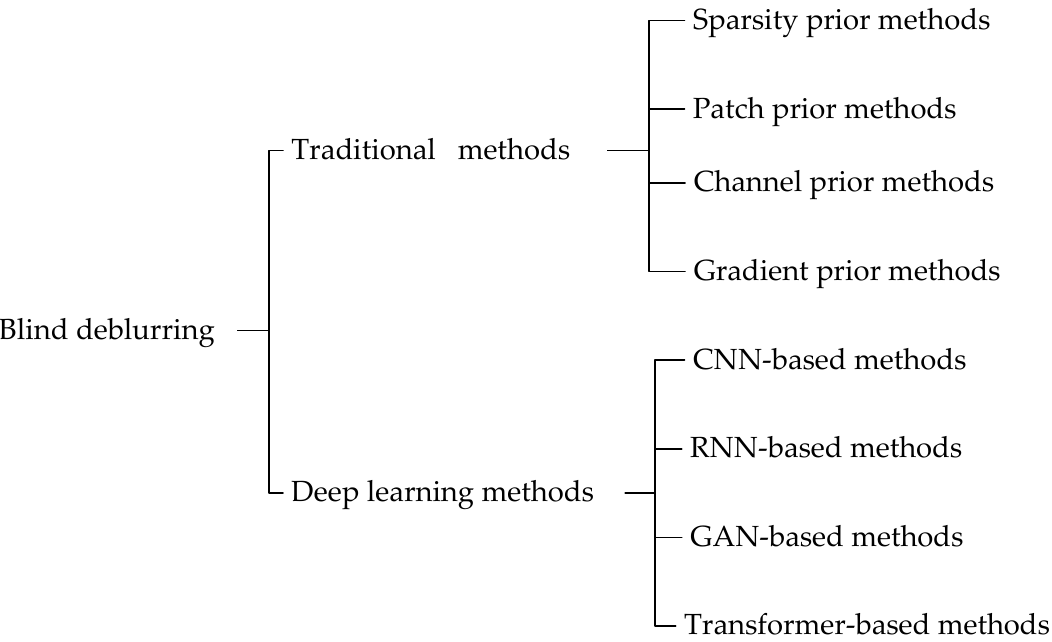}
	\caption{An overview of traditional and deep learning methods for blind motion deblurring.}
	\label{Fig.traditional}
\end{figure}

{Deblurring represents an ill-posed problem that cannot be directly solved.
The traditional blind deblurring methods~\cite{levin2006blind,yang2019variational,fergus2006removing,dong2017blind} start with the blur degradation model, employing mathematical modeling or convolutional networks to predict the blur kernel. Then they utilize the estimated blur kernel to restore the image through deconvolution.}
These approaches rely on prior knowledge of both the image and the blur kernel to address the ill-posedness in deblurring. 
As shown in the traditional method on \figurename~\ref{Fig.traditional}, the commonly used prior algorithms are sparsity prior~\cite{krishnan2011blind,xu2013unnatural,levin2011efficient}, patch prior~\cite{sun2013edge,guo2018image,tang2018blind}, channel prior~\cite{pan2017deblurring,ge2021blind,cai2020dark}, gradient prior~\cite{levin2009understanding,pan2014deblurring,shao2020gradient} and so on.
These traditional methods heavily rely on blur kernel estimation, and the effectiveness of the deblurring process directly depends on the accuracy of the kernel estimation. However, in complex real-world scenarios, the blur kernel in the blurring process tends to be complex, variable, and difficult to parameterize, causing the traditional methods to often perform inadequately~\cite{sheng2019depth}.
{Deblurring methods based on deep learning can learn the nonlinear mapping of blurred images to clear images end-to-end. The effect of inaccurate blur kernel estimation on the deblurring is avoided. Deep learning methods tend to have better deblurring results in dealing with non-uniform blurring problems in the real world.}

{In recent years, deep learning technology has developed rapidly and has been widely applied in image processing~\cite{zhang2012single,zhu2022fffn,jiao2020blur,10443067,dong2020deep,pan2023deep,Li2023self}.}
Deep learning-based methods have powerful feature learning and mapping capabilities, allowing them to learn complex patterns of blur removal from extensive datasets~\cite{gu2020blur}.
Consequently, significant progress has been made in image deblurring\cite{zhao2022rethinking,purohit2021spatially,chen2022simple,yang2022motion,nimisha2017blur}.

\begin{figure}[t]
	\centering 
	\includegraphics[width=13cm]{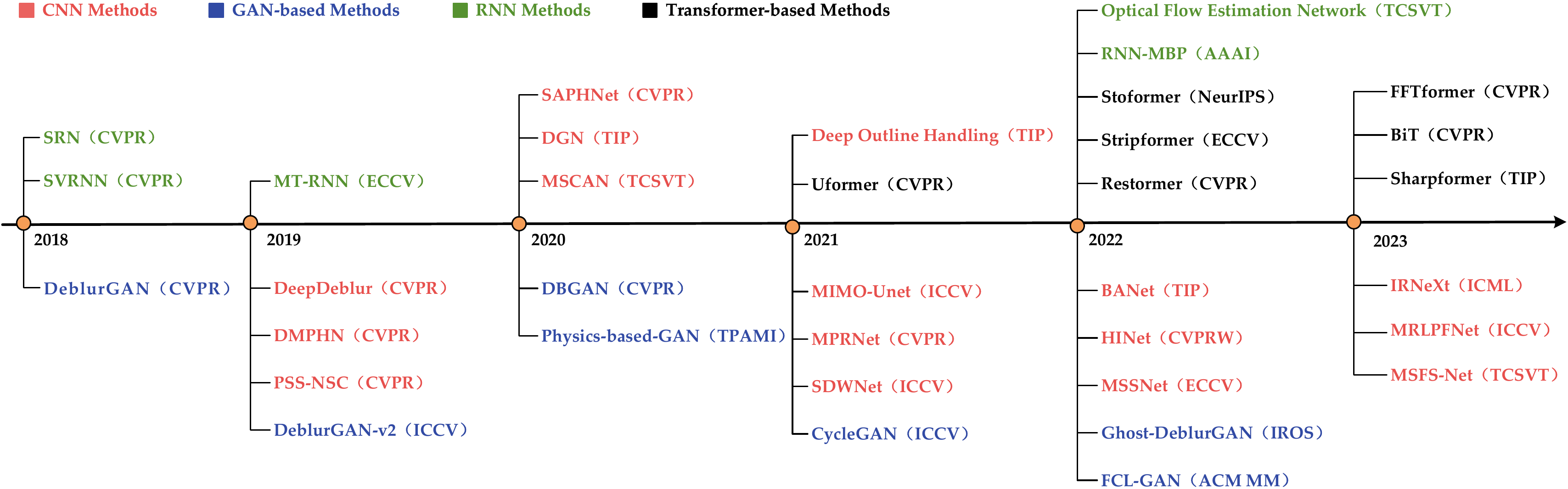}
	\caption{{An overview of deep learning methods for blind motion deblurring.}}
	\label{Fig.time_sequence}
\end{figure}

{As depicted in \figurename~\ref{Fig.time_sequence}, existing blind motion deblurring methods can be categorized based on different backbone networks, namely: convolutional neural networks (CNN)~\cite{mao2016image,pan2023cascaded,dong2021deep}, generative adversarial networks (GAN)~\cite{goodfellow2014generative,lu2022image}, recurrent neural networks (RNN)~\cite{zhang2018dynamic}, and Transformer networks~\cite{zhao2023wtransu}.}
The deep learning methods can adaptively learn the blur features according to the training data.
They can directly generate sharp images from blurred ones, which improves the effect of image deblurring. 
Compared to the traditional methods, the deep learning methods~\cite{li2021perceptual,yuan2020efficient,noroozi2017motion} demonstrate improved robustness in real-world scenarios and possess broader applicability.

In this paper, we focus on the application of deep learning in blind motion deblurring.
Overall, the main contributions of this paper are as follows:
\begin{itemize}
	\item  In this review, we comprehensively overview the research progress of deep learning in blind motion deblurring, encompassing the causes of motion blur, blurred image datasets, evaluation metrics for image quality, and methods developed over the past six years.
	
	\item Specifically, we introduce a classification framework for blind motion deblurring tasks based on backbone networks, categorizing existing methods into four classes: CNN-based, RNN-based, GAN-based, and Transformer-based approaches.
	
	\item For each type of blind motion deblurring method, we provide an in-depth summary of their fundamental principles, analyzing the advantages and limitations of each approach. Qualitative and quantitative experimental results on four widely used datasets further showcase the performance differences among different blind motion deblurring methods.
	
	\item Finally, we conduct a prospective analysis and summary of current research challenges and potential future directions in blind motion deblurring. This review aims to offer researchers and practitioners in the field a comprehensive understanding and a valuable reference for further advancements.
\end{itemize}

The subsequent sections of this paper are structured as follows: We first summarized the classification and research status of image blind motion deblurring methods in Sec.~\ref{sec.deeplearning}.
Then, We briefly introduced the public datasets and evaluation metrics in Sec.~\ref{sec.dataset}.
Next, Sec.~\ref{sec.methods} describes the basic principles of different types of methods in detail, and summarizes and compares the advancements and shortcomings of different methods. Qualitative and quantitative experimental results on four common datasets further compare the performance of different methods.
Subsequently, in Sec.~\ref{sec.challenge}, we provide a prospective analysis of current research challenges and potential future directions in blind motion deblurring.
Finally, in Sec.~\ref{sec.conclusion}, we provide a summary of our paper.

\section{Deep Learning Methods for Blind Motion Deblurring}
\label{sec.deeplearning}
As shown in \figurename~\ref{Fig.time_sequence}, the foundational framework of deep learning-based image deblurring methods can be divided into four categories: CNN-based,  RNN-based, GAN-based, and Transformer-based image deblurring methods. We discuss these methods in the following sections.

\subsection{CNN-based Blind Motion Deblurring Methods}
\begin{figure}[ht]
	%	\centering  % ????
	\includegraphics[width=12 cm]{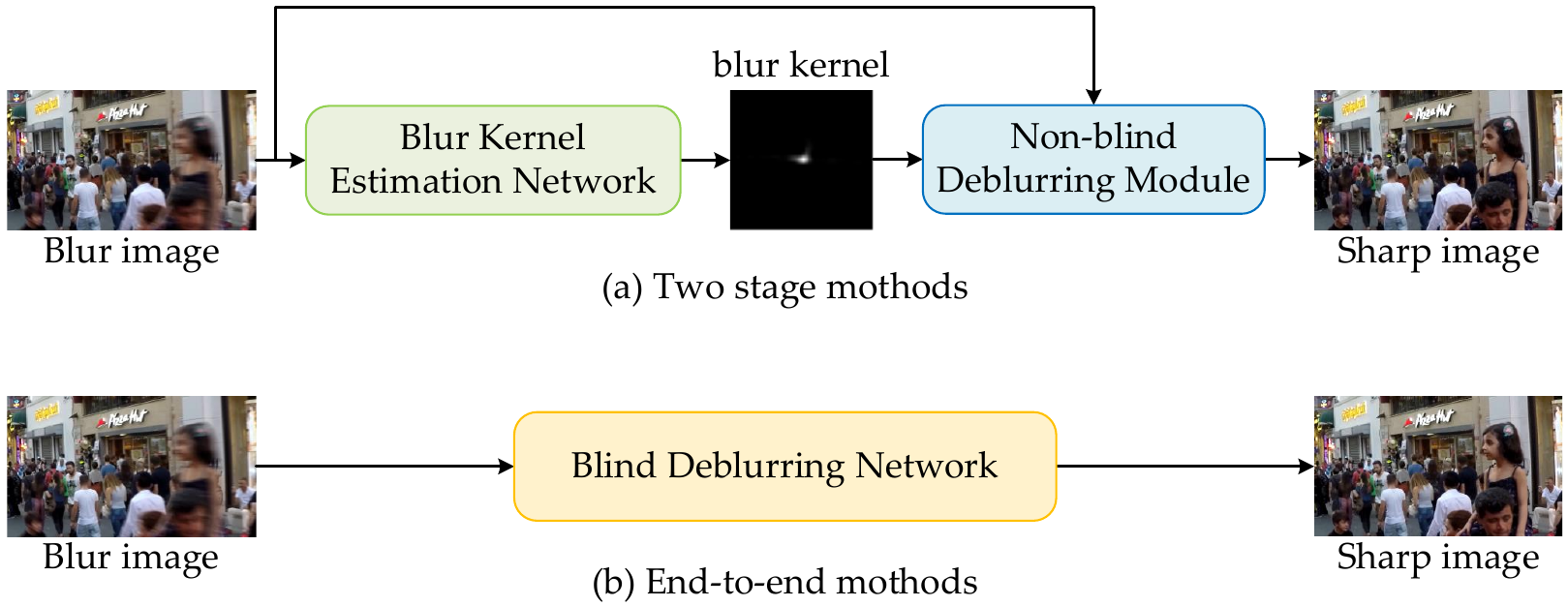}
	\caption{Two CNN-based blind motion deblurring frameworks.}
	\label{Two_stage_CNN}
\end{figure}

CNN-based image blind deblurring methods can be divided into two categories according to the process of image deblurring. 
One is the early two-stage image deblurring framework shown in \figurename~\ref{Two_stage_CNN} (a), and the other is the end-to-end image deblurring framework with the better performance shown in \figurename~\ref{Two_stage_CNN} (b).

\subsubsection{Two-stage CNN-based Blind Motion Deblurring Methods}
Early image blind deblurring methods are mainly for deblurring a single blur kernel image. The process of image deblurring was divided into two stages, as shown in Figure 4 (a).  The first stage is estimating the blur kernel through a neural network. Then, the estimated blur kernel is used for deconvolution or inverse filtering operations on the blurred image to achieve the effect of image deblurring. 

 Schuler et al.~\cite{schuler2015learning} trained a blind image deconvolution network to deblur the image. This network primarily consists of three parts: a feature extraction module, a blur kernel estimation module, and an image estimation module. The estimated blur kernel from the blur kernel estimation module is used to iteratively update the deblurred image. Sun et al.~\cite{sun2015learning} proposed to estimate the heterogeneous motion blur through the CNN model and then deconvolute the blurred images, which can effectively estimate the spatially varying motion kernel and achieve a better effect of removing motion blur.
To address the issue of the non-uniform blur, Cronje et al.~\cite{cronje2015deep} proposed a CNN-based image deblurring method. This method trained two CNN networks, one for estimating the vector of non-uniform motion and the other for image deconvolution. It accurately estimated non-uniform blur and enhanced the effectiveness of image deblurring. Ayan Chakrabarti et al.~\cite{chakrabarti2016neural} trained a neural network for estimating the blur kernel of blur images. Using the estimated single global blur kernel, they performed deconvolution operation on the blur images to obtain deblurred images. Gong et al.~\cite{gong2017motion} trained a fully convolutional neural network for estimating image motion flow. Then they preformed non-blind deconvolution to recover clear images from the estimated motion flow. This method avoids the iterative process of underlying image priors. Xu et al.~\cite{xu2017motion} proposed a deep convolutional neural network to extract sharp edges from blur images for kernel estimation, which does not require any coarse-to-fine strategy or edge selection, thus greatly simplifying the kernel estimation and reducing the computational amount. Kaufman et al.~\cite{kaufman2020deblurring} analyze the blurring kernel by applying convolutional layers and pooling operations to extract features at multiple scales. Subsequently, the estimated kernel, along with the input image, is fed into a synthetic network to perform non-blind deblurring.

These two-stage image deblurring methods rely too much on the blur kernel estimation of the first stage, and the estimated blur kernel quality is directly related to the effect of image deblurring. In reality, the blur is mostly uneven, and the size and direction of the blur are also uncertain. So in the real scenes, this method can not remove complex real blur very well.

\subsubsection{End-to-end CNN-based Blind Motion Deblurring Methods}
\begin{figure}[t]
	
	\centering  % ????
	\includegraphics[width=13cm]{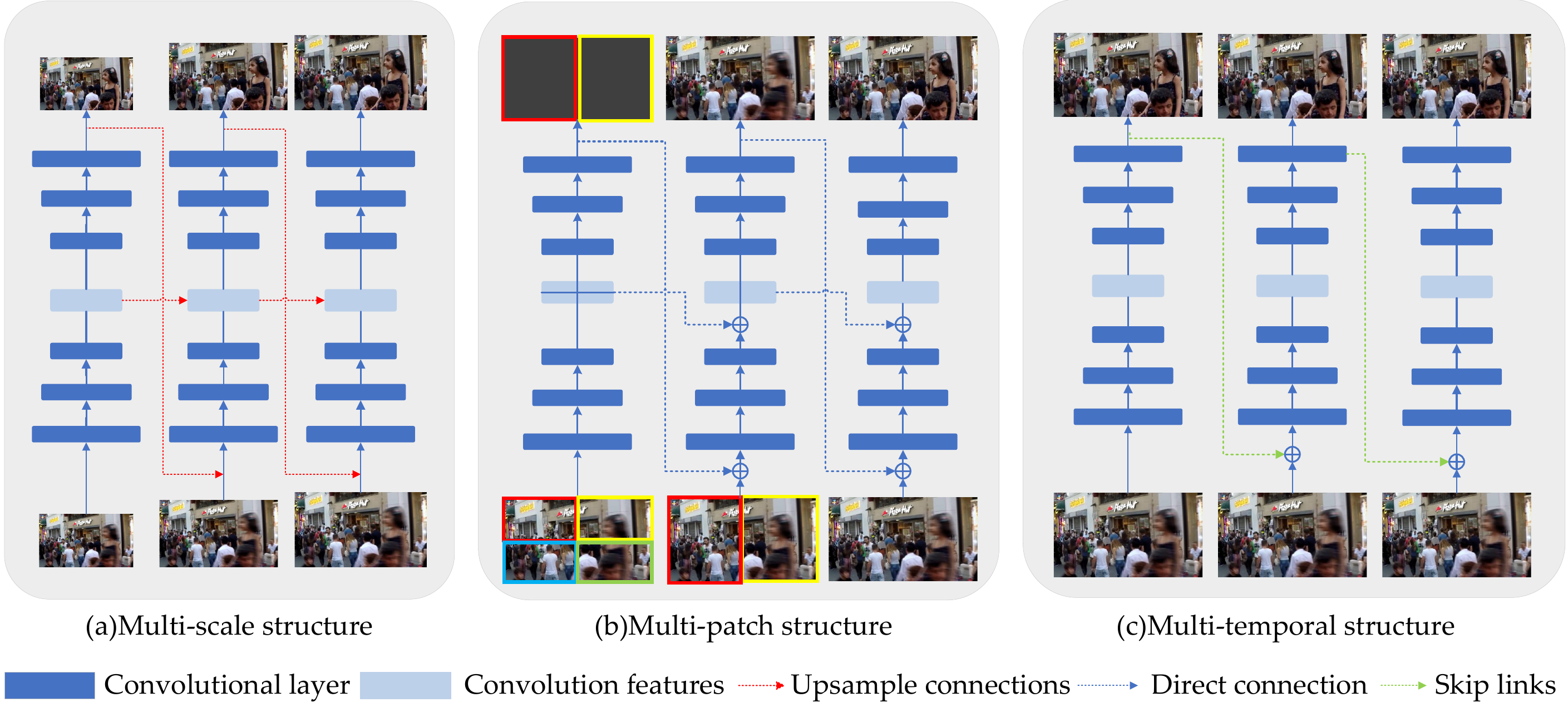}
	\caption{Different end-to-end CNN-based blind motion deblurring network architectures.}
	\label{CNN}
\end{figure}

The end-to-end image deblurring method generates a clear image directly from the input blurred image. {It utilizes neural networks to learn complex feature mapping relationships for better restoration of clear images~\cite{pan2021deep}.} End-to-end image deblurring algorithms have made significant progress since Xu et al.~\cite{xu2014deep} first implemented end-to-end motion blur image restoration using CNNs. As shown in \figurename~\ref{CNN}, CNN-based end-to-end image deblurring methods can be categorized into three types: multi-scale, multi-patch, and multi-temporal structures.

\textbf{Multi-scale Structure}:As shown in \figurename~\ref{CNN} (a), the multi-scale model extracts multi-scale feature information from blurry images. Based on this extracted information, it progressively restores the image from coarse to fine levels\cite{gao2023efficient}. In order to better handle non-uniform blur in dynamic scenes, Nah et al.~\cite{nah2017deep}proposed a multi-scale convolutional neural network for end-to-end restoration of blurred images. Following a similar approach, many subsequent researchers have improved on the multi-scale strategy to obtain better results~\cite{purohit2020region,tao2018scale,vitoria2022event,hemanth2020dynamic}. However, its limitation lies in scaling the blurred image to a lower resolution, often resulting in the loss of edge information. Overlaying multi-scale input images onto sub-networks gradually enhances image clarity from lower-level to higher-level sub-networks, inevitably incurring higher computational costs. {Dong et al.~\cite{dong2023multi} proposed a simple and effective multi-scale residual low-pass filtering network, which can adaptively explore the spatially varying properties of the global context and effectively model both high-frequency and low-frequency information.} To reduce the spatio-temporal complexity of employing a multi-scale strategy network, Cho et al.~\cite{cho2021rethinking} devised a network architecture that starts from coarse to fine, proposing a multi-input multi-output U-Net network. This approach allows for a better execution of deblurring from coarse to fine.
{To address the oversight of image frequency distinction based on multi-scale methods, Zhang et al.~\cite{zhang2023multi} proposed a multi-scale frequency separation network. By combining multi-scale strategies with frequency separation, it enhances the restoration of details in blurry images.}

\textbf{Multi-patch Structure}: As shown in \figurename~\ref{CNN} (b), the multi-patch structure segments the blurred input into multiple patches for prediction. 
Inspired by spatial pyramid matching, Zhang et al.~\cite{zhang2019deep} proposed a deep hierarchical end-to-end CNN model called DMPHN. This model used multiple patches for a fine-to-coarse operation to handle blurry images. To remove spatially varying blur, Suin et al.~\cite{suin2020spatially} proposed a patch hierarchical attention architecture. It significantly enhances deblurring effects through effective pixel-wise adaptation and feature attention. Building upon DMPHN, Zhang et al.~\cite{zhang2023event} proposed a self-supervised event-guided deep hierarchical multi-patch network, which handles blurry images through a layered representation from fine to coarse. By stacking patch networks to increase network depth, this model enhanced deblurring performance. However, when segmenting input or features, discontinuities in contextual information may arise, potentially generating artifacts and leading to suboptimal image deblurring in dynamic scene contexts~\cite{zamir2021multi}.

\textbf{Multi-temporal Structure}: As shown in \figurename~\ref{CNN} (c), the multi-temporal structure gradually eliminates non-uniform blur over multiple iterations. The multi-temporal framework~\cite{park2020multi} was introduced to handle non-uniform blur and exhibited significant performance improvements. However, rigid progressive training and inference processes might not adapt well to images with varying degrees of blur in different regions. Subsequent researchers have proposed numerous time-series-based deblurring methods, learning from different temporal scales of motion blur to achieve enhanced deblurring effects~\cite{zhang2023generalizing,zhang2023crosszoom}.

\subsection{RNN-based Blind Motion Deblurring Methods}

\begin{figure}[t]
	
	\centering  % ????
	\includegraphics[width=13cm]{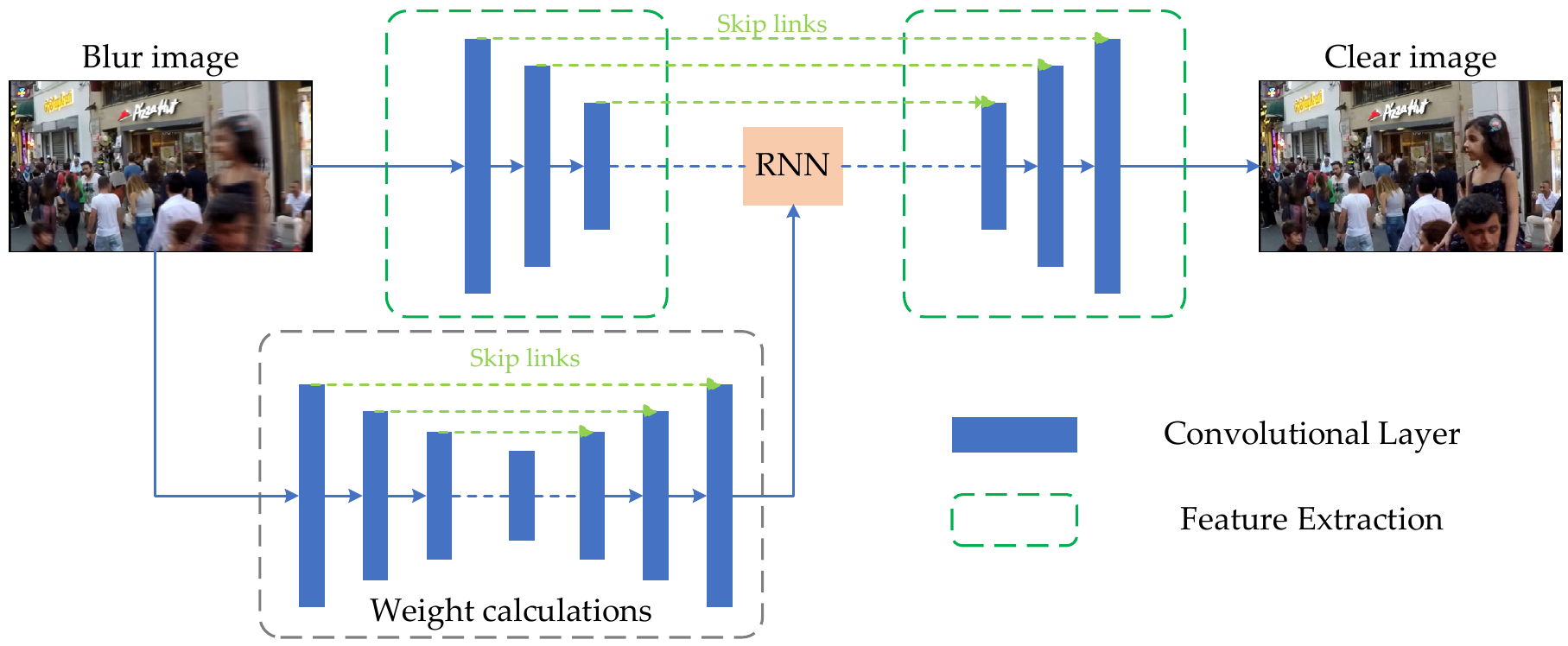}
	\caption{A spatial variant RNN image deblurring method.}
	\label{SVRNN}
\end{figure}

Recurrent Neural Networks (RNNs) are commonly used for handling sequential data such as text, language, and time series, demonstrating exceptional performance when dealing with data that exhibit temporal or sequential patterns~\cite{liping2022overview}. Due to their memory-based neural network architecture, RNNs have found applications in image processing as well. For instance, Zhang et al.~\cite{zhang2018dynamic} proposed an end-to-end spatially variant RNN network for deblurring in dynamic scenes. As illustrated in \figurename~\ref{SVRNN}, the weights of RNNs are learned by deep CNNs. They analyzed the relationship between spatially variant RNNs and the deconvolution process, demonstrating the ability of spatially variant RNNs to simulate the deblurring process. Models trained using the proposed RNNs are notably smaller and faster. To better utilize the properties of spatially varying RNNs, Ren et al.~\cite{ren2021deblurring} extend a unidirectionally connected one-dimensional spatially varying RNN to a two-dimensional spatial RNN with three-way connectivity. 2D RNNs can capture a larger receptive field and learn denser propagation between neighboring pixels compared to 1D RNNs, further improving deblurring performance. Tao et al.~\cite{ tao2018scale} proposed a multi-scale deep neural network for single image deblurring based on convolutional long short-term memory (Conv LSTM), effectively utilizing previous frames and image features. Dongwon Park et al.~\cite{park2020multi} developed a multi-temporal recurrent neural network (MT-RNN) based on recursive feature mapping for blind single-image deblurring, progressively deblurring a single image in an iterative process. {Zhang et al.~\cite{dong2023multi} proposed an effective dynamic scene deblurring method based on optical flow and spatial variable RNN, which utilizes optical flow to guide the spatially variable RNN to remove blur.}
In practical applications, CNNs and other models are often combined with RNNs to compensate for certain limitations in image processing, such as sequence length restrictions and computational complexity.

\subsection{GAN-based Blind Motion Deblurring Methods}
\begin{figure}[t]
	
	\centering  % ????
	\includegraphics[width=13cm]{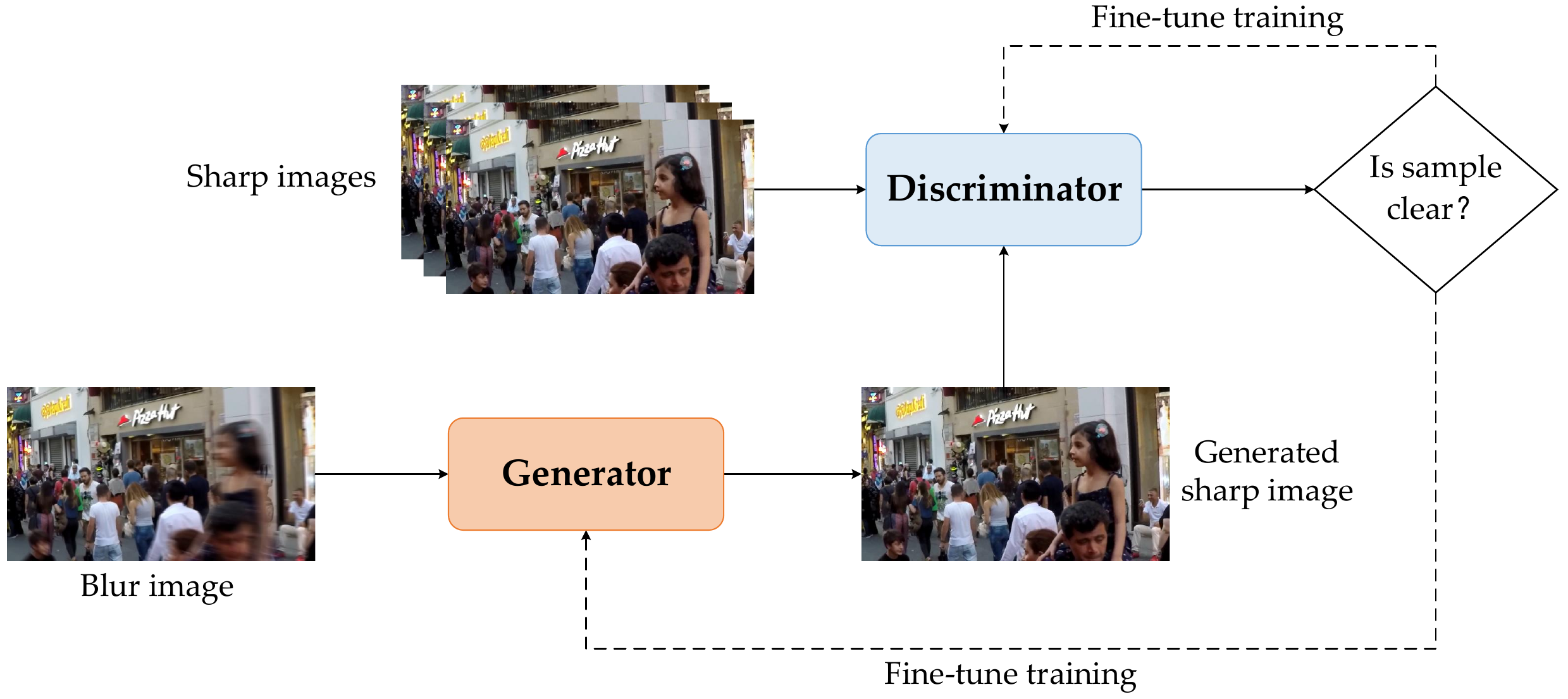}
	\caption{GAN-based deblurring algorithm flow.}
	\label{GAN}
\end{figure}

With GANs~\cite{goodfellow2014generative} showing promising results in computer vision tasks, they have also been applied to image deblurring tasks. The basic process of image deblurring based on GANs is illustrated in \figurename~\ref{GAN}. The generator learns to restore clear images from blurry ones, while the discriminator assesses whether the generated clear images are real, providing feedback to adjust the training of both the generator and discriminator. {GAN-based methods can be divided into two categories, supervised and unsupervised models, based on different training methods and data dependencies.}

\textbf{Supervised methods:}
{
	Supervised methods use blur-sharp image pairs as supervisory signals for neural network training to guide model updating and learning. Supervised motion deblurring can more accurately learn the nonlinear mapping between blurred and sharp images, better removing blur and restoring details.}
Ramakrishnan et al.~\cite{ramakrishnan2017deep} proposed a depth filter based on GAN. For batter restoration performance, it combines global jump connections and dense structure. Additionally, the model omits the blur kernel estimation process, significantly reducing the testing time required for practical applications. Lin et al.~\cite{lin2019tell} proposed an adversarial blurred region mining and refinement method. This method can drive CGAN~\cite{mirza2014conditional} to learn real sharp image distributions globally and locally for blind image deblurring. Based on GANs, O.Kupyn et al.~\cite{kupyn2018deblurgan} proposed an end-to-end image deblurring network termed deblurGAN. DeblurGAN was trained based on WGAN~\cite{arjovsky2017wasserstein} with gradient penalty~\cite{gulrajani2017improved} and perceptual loss~\cite{johnson2016perceptual}, achieving notable deblurring effects. Based on DeblurGAN, O. Kupyn et al.~\cite{kupyn2019deblurgan} proposed DeblurGAN-v2. This model incorporates a feature pyramid network within the generator. The discriminator employs a relativistic discriminator and uses least squares loss to assess the generated images on both global and local scales. Moreover, recognizing the impact of backbone network selection on deblurring quality and efficiency, DeblurGAN-v2 utilizes an embedded architecture that allows the use of different backbone networks for feature propagation. Compared to DeblurGAN, DeblurGAN-v2 demonstrates significant improvements in both performance and efficiency. Peng et al.~\cite{peng2023mnd} also proposed the MobileNetDeblur-GAN (MND-GAN) deblurring algorithm based on the DeblurGAN algorithm. This algorithm utilizes MobileNet as a feature extractor to build a feature pyramid network. Additionally, it employs a multi-scale discriminator to jointly train the generator, reducing the computational cost of the generator while ensuring deblurring quality. 

To further enhance image deblurring performance, Zhang et al.~\cite{zheng2019edge} proposed a multi-scale deblurring network based on GANs. The model first restores the edge information of the blurred image and then utilizes clear edges to guide a multi-scale network for image deblurring. Starting from the blurry images themselves, Zhang et al.~\cite{zhang2020deblurring} proposed an image deblurring approach that combines two GAN models: a BGAN subnetwork for learning the image blurring process to generate motion-blurred images and a DBGAN for learning the restoration of motion-blurred images. By learning how to blur images, this method can better remove blurriness.
{Pan et al.~\cite{pan2021physics} applied the physical model to generative adversarial networks for image restoration, and proposed a GAN model constrained by the physical model. The physical model is derived from the image formation process of the underlying problem.
Considering the practical application of image deblurring algorithms in real-world scenarios, Liu et al.~\cite{liu2022application} proposed a lightweight generative adversarial network termed Ghost-DeblurGAN. While this algorithm may not reach the advanced levels of deblurring, its advantage lies in its small model size and short inference time.
}

\textbf{Unsupervised methods:}
Supervised learning relies on a large amount of training data, but in reality, there are few real blurred-clear image pairs available for collection. Madam et al.~\cite{nimisha2018unsupervised} proposed an unsupervised motion image deblurring method based on GANs. The model employs GANs to learn the mapping relationship from blurred images to clear images. To prevent mode collapse and maintain consistency between deblurred and clear image gradients, the model incorporates a deblurring CNN module and a gradient module to constrain the solution space and guide the deblurring process. GAN-based methods may encounter issues during training such as model collapse and gradient disappearance\cite{mescheder2018training}, potentially resulting in artifacts and loss of details in the reconstructed images. Building upon the cycle-consistent generative adversarial network (CycleGAN)~\cite{zhu2017unpaired}, Wen et al.~\cite{wen2021structure} proposed an unsupervised image deblurring method using a multi-adversarial optimization CycleGAN to address artifacts in high-resolution image generation. This method iteratively generates clear high-resolution images through multiple adversarial constraints. Considering the significant role of edge restoration in the structure of deblurring blurry images, this method introduces a structural-aware strategy. It utilizes edge information to guide the network in deblurring, enhancing the network's ability to recover image details. Zhao et al.~\cite{zhao2023real} proposed a contrast-guided multi-attention CycleGAN unsupervised algorithm for motion deblurring. It improved the deblurring performance of unsupervised algorithms. Lu et al.~\cite{lu2019uid} proposed an unsupervised method for single image deblurring without pairing training images termed UID-GAN. The method introduces a disentangled framework to split the content and blur features of blurred images, which improves the deblurring performance.
{
Due to the current drawbacks of unsupervised learning, such as large model sizes, lengthy inference times, and strict constraints on image resolution and domains, Zhao et al.~\cite{zhao2022fcl} proposed a lightweight CycleGAN (FCL-GAN) with frequency domain contrast loss constraints. 
FCL-GAN designed lightweight domain transformation units and parameter-free frequency domain contrast units to ensure the algorithm's "lightweight" and "real-time" operation, overcoming the limitations of the generally large models and complex inference in unsupervised methods. 
%Additionally, this model enhanced deblurring performance by introducing frequency-domain contrastive learning.
 }

\subsection{Transformer-based Blind Motion Deblurring Methods}
In recent years, models based on transformers have demonstrated tremendous potential in visual tasks~\cite{lin2021eapt,zhao2023comprehensive}. The Transformer, renowned for its powerful attention mechanism, utilizes self-attention (SA) to capture long-range dependencies among patches in image sequences and adaptability to given input content~\cite{vaswani2017attention,liu2021swin,raghu2021vision}.
When applied in image deblurring, it can achieve significant improvements in deblurring effectiveness.
{
	The Transformer model itself possesses the capability to handle information of varying scales and ranges, allowing it to flexibly adapt to different types of blur problems. According to the specific image blurring situation and deblurring requirements the transformer-based deblurring methods are divided into two categories: global and local deblurring. 
	}

%	The Transformer model itself possesses the capability to handle information of varying scales and ranges, allowing it to flexibly adapt to different types of blur problems. According to the specific image blurring situation and deblurring requirements the transformer-based deblurring methods are divided into two categories: global and local deblurring. For tasks requiring local deblurring, leveraging the lower-level feature extraction layers of the Transformer facilitates the capture of detailed information within specific regions of the image. Then, through higher-level feature extraction layers, restoration of locally blurred portions can be achieved. In tasks involving global deblurring, the Transformer can capture long-range dependencies between different regions, thereby aiding in the modeling and handling of global information. 	 

\begin{figure}[t]

	\centering  % ????
	\includegraphics[width=13cm]{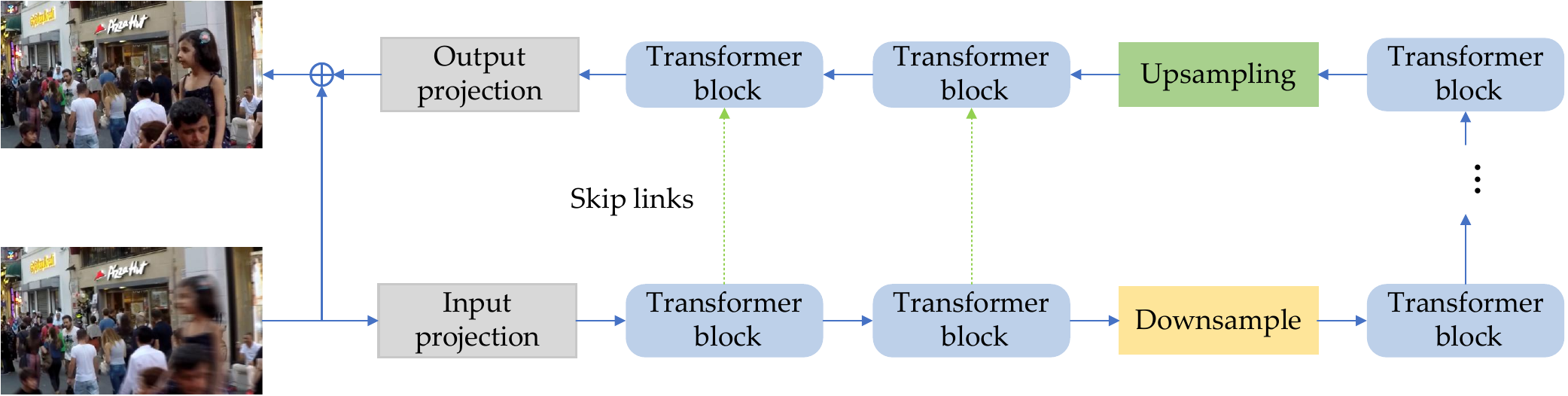}
	\caption{The Transformer-based deblurring network architecture.}
	\label{Uformer}
\end{figure}

\textbf{Global deblurring methods:}
{Global deblurring methods are mostly applied to the blurring of the whole image produced by camera shake. Global deblurring methods do not distinguish between sharp and blurred regions and consider the whole image as a blurred whole. The same deblurring process is done for blurred and sharp regions.}
Wang et al.~\cite{wang2022uformer} proposed an efficient and effective Transformer-based image restoration architecture called U-former by embedding local enhancement window Transformer blocks into a U-shaped structure~\cite{ronneberger2015u}. They inserted learnable multi-scale restoration modulators into their decoder to enhance the quality of image restoration. The basic architecture is depicted in \figurename~\ref{Uformer}. Although Uformer has shown promising performance in image restoration tasks, the heavy computation involved in self-attention restricts the application of the transformer within the encoding-decoding framework to limited depths. This inevitably affects the effectiveness of image restoration. 
Building upon U-former, Feng et al.~\cite{10155416} proposed a deep encoder-decoder image restoration network called U2-former, with Transformer as the core operation. This model employs two nested U-shaped structures to facilitate feature mapping at different scales, resulting in an enhanced image deblurring effect. To optimize computational efficiency, the model utilizes a feature filtering mechanism to discard low-quality features, thereby reducing unnecessary computations.

To effectively eliminate non-uniform blur, Tsai et al.~\cite{tsai2022stripformer} proposed an efficient Transformer-based image deblurring algorithm termed Stripformer. This method reweights image features by computing intra-band and inter-band attention to address blur of different sizes and orientations. It achieves notable deblurring results without requiring extensive training data, enhancing the removal of non-uniform blur.
{
	Zamir et al.~\cite{zamir2022restormer} proposed a Transformer-based image restoration network termed Restormer. Restormer employs a multi-head attention mechanism for feature fusion to handle high-resolution images in image restoration tasks. Additionally, this method utilizes multi-scale local-global representation learning to effectively utilize distant contextual information. It simultaneously reduces computations while enhancing the effectiveness of image restoration.
	Kong et al.~\cite{kong2023efficient} designed an effective frequency-domain self-attention solver in the decoder, reducing spatial complexity and computational demands through element-wise operations. Additionally, they employed a discriminative frequency-based feedforward network (DFFN) to discern which low-frequency and high-frequency feature information should be retained for potential sharp image restoration.
}
Recently, Zhong et al.~\cite{zhong2023blur} proposed a Blur Interpolation Transformer (BiT) to reveal potential temporal correlations in blurry encodings. Based on multi-scale residual Swin Transformer blocks, they introduced dual-ended temporal supervision and a time-symmetric integration strategy to generate effective time-varying motion-rendered features, effectively generating clear image sequences from blurry inputs. Zou et al.~\cite{zou2023edgeformer} explored the significant roles of edges and textures in image deblurring and proposed a method called Edgeformer. This method utilizes edge information to enhance the correlation between pixels. 
By employing an edge-enhanced feed-forward network, it distinguishes between edge and texture features, resulting in improved image deblurring. To fully leverage both local and non-local features in images, Wu et al.~\cite{wu2023hierarchical} introduced the hierarchical pyramid attention transformer (HPAT) network, based on a transformer-based multi-level network structure. It employs hierarchical attention fusion modules (HAFM) to extract global contextual information, enhancing the feature representation capability. Using a fusion feed-forward neural network (F3N) for feature fusion has improved the deblurring effectiveness. Liang et al.~\cite{liang2023image} used deep features extracted from pre-trained visual transformers (ViT)~\cite{dosovitskiy2020image} to encourage the recovered image to become clear without sacrificing quantitative performance. They also proposed local MAE perceptual loss and global distribution perceptual loss to guide image deblurring.

\textbf{Local deblurring methods:}
The existing image deblurring methods mainly focus on global deblurring, which may affect the clarity of locally blurred images. 
{ The local deblurring method mainly focuses on the situation where the foreground and background are moving relative to each other. The foreground is blurred while the background is sharp. Local deblurring methods will separate the blurred foreground region by taking mask and other operations to ensure that there are no adverse effects on the sharp regions during the deblurring process.}
Li et al.~\cite{li2023adaptive} proposed a local dynamic motion deblurring transformation network (LDM-ViT) based on adaptive window pruning transformer blocks (AdaWPT). In order to focus the network attention on localized blur regions, AdaWPT removes unnecessary windows. Only the blurred regions are allowed to participate in the deblurring process, reducing the amount of unnecessary computation. Additionally, using labeled local blur image datasets to train the predictor can be able to remove local motion blur effectively.
Yan et al.~\cite{yan2023sharpformer} proposed a Transformer based deblurring network for dynamic scene images termed SharpFormer. To effectively model the detailed features of the image, the model uses locality preserving Transformer (LTransformer) blocks to learn the local features. In addition, the model uses dynamic convolutional blocks to adaptively deal with the non-uniform blurring of the input images, improving the quality of the deblurred image.

\subsection{Advantages and limitations of the four architectures.}

In image deblurring, the four architectures CNN, RNN, GAN, and Transformer have their own advantages and limitations as shown in~\tablename~\ref{tab:advantages} :

CNN is widely used in image processing to capture local features and spatial information~\cite{li2018learning,liu2021efficient}. When trained on large datasets, deblurring methods based on CNN perform high efficiency and generalization ability. 
CNN demonstrates adaptability to different types of blurry scenarios, enabling it to handle various forms of blur such as motion blur and defocus blur. The CNN structure is simple and suitable for image denoising and deblurring tasks~\cite{tian2020deep}. However, CNN performance typically relies on extensive training data, requiring a substantial number of clear and blurry image pairs for effective training. Additionally, CNN may be constrained by a fixed-size receptive field, potentially resulting in suboptimal performance in image deblurring tasks involving global information or long-range dependencies. In order to solve the problem of limited receptive field, the more widely used method is to use the dilated convolution.

RNN can effectively capture temporal information in images, aiding in handling blurs with time dependencies, such as motion blur. However, there may be gradient vanishing or explosion problems when dealing with long-term dependencies~\cite{rusch2021unicornn}. 
Moreover, for image deblurring tasks, RNN is not well suited to capture spatial information~\cite{ren2021deblurring}. Therefore, RNNs are generally combined with other structures to accomplish image deblurring tasks.

GAN enhances the realism of image generation through adversarial training, providing a significant advantage in deblurring tasks. 
The GAN generator can produce diverse outputs, helping to deal with different scenarios and different types of blurring, thereby improving the model's generalization ability. 
{In contrast to CNN and Transformer methods, the generator training of GAN does not depend entirely on data samples, and its gradient update information comes from the discriminator. GAN can learn the task of image deblurring through unsupervised learning from a large amount of unlabeled data. Image motion deblurring methods based on GAN improve the accuracy of parameter fitting based on the principles of game theory. }
However, the training process may be unstable, requiring a balance between generator and discriminator training~\cite{cui2021efficient}. There may also be issues such as pattern crashes or non-convergence of training patterns.

Transformer is able to capture global information and solve the remote spatial dependence problem~\cite{yan2023sharpformer}, and has advantages in processing for some image tasks that require long-distance dependence. Transformer can capture the global information of input images, aiding in understanding the overall blurry structure and image content for better restoration of clear images. The introduction of positional encoding in Transformer enables effective handling of spatial relationships in images, making the model more targeted in dealing with blur. The self-attention mechanism of Transformer allows it to dynamically adjust its focus on different regions of the image, helping to address varying degrees of blur.
However, the image deblurring task involves the processing of a large number of pixels and has a high computational cost. Compared to traditional CNN structures, Transformer models typically have a larger parameter count~\cite{zhao2023comprehensive}, which may result in higher training and inference costs.

%{
%In practice, these methods can be used independently or in combination to fully leverage their respective strengths in addressing image deblurring problems. There have been some methods that use two of CNN, RNN, GAN, and Transformer in combination to take advantage of their respective strengths and achieve better deblurring results. The SVRNN~\cite{zhang2018dynamic} method utilizes RNN to learn inter-pixel weights and CNN to remove blur. The CTMSS~\cite{zhao2022rethinking} method combines CNN and transformer to fully utilize their respective advantages. Zhao et al.~\cite{zhao2023wtransu} proposed a novel network combined with transformer, CNN and multi-scale technologies to better improve the deblurring performance.}

\begin{landscape}

	\begin{longtable}[c]{p{2.2cm}p{0.8cm}cp{5cm}p{4.5cm}p{4.5cm}}

		\caption{
			%			Overview of deep learning based deblurring methods.
			{A review of deep learning based deblurring methods, including publication information, overview, advantages and limitations.}
		}
		\tiny
		\tiny
		\label{tab:advantages}\\		\toprule
		\textbf{Methods}           & \textbf{Pub.}      & \textbf{Category}        & \textbf{Overview}      & \textbf{Advantages}     & \textbf{Limitations}     \\* \midrule
		\endhead
		\bottomrule
		\endfoot
		\endlastfoot
		%
		%		\hline
		%		Model methods           & Time      & Type        & Deblurring Methods      & Advantage     & Limitations  \\ 
		%		\hline 
		%		\endfirsthead 
		%		\multicolumn{2}{c}{{Table\thetable{} (xu)}} \\
		%		\hline
		%		Model methods           & Time      & Type        & Deblurring Methods      & Advantage     & Limitations \\ 
		%		\hline 
		%		\endhead
		DeepDeblur~\cite{nah2017deep}     & CVPR 2017  & CNN         & Removing complex motion blur from coarse to fine using a multi-scale recurrent network                                                                                                                                                                                                                                             & End-to-end removal of complex motion blur without the need for kernel estimation                                                                                                                          & The mode has a large number of parameters,making training difficult     \\
		
		DMPHN~\cite{zhang2019deep}           & CVPR 2019  & CNN         & Estimated by segmenting the image into multiple blocks and then removing blur at different scales                  & Overcoming the limitations of depth deblurring models on superposition depths              & Feature segmentation may lead to discontinuous contextual information                             \\
		
		PSS-NSC~\cite{gao2019dynamic}           & CVPR 2019  & CNN         &Multi-scale CNN deblurring networks are analyzed and a parameter-selective sharing scheme is proposed                   &Selective parameter sharing can improve the learning efficiency and generalization of the model              &The introduction of new network structures and hopping connection mechanisms may have increased computational costs                 \\
		
		DGN~\cite{li2020dynamic}           & TIP 2020  & CNN         &An effective depth-guided network consisting of an image deblurring branch and a depth refinement branch is proposed        &The use of depth maps allows for a better understanding of scene changes and facilitates the use of depth information     &Deblurring quality depends on the accuracy and quality of the depth map      \\
		
		MSCAN~\cite{wan2020deep}           & TSCVT 2020  & CNN         &A multi-scale channel attention network is proposed for more robust feature representation and higher quality image recovery        &Further taking advantage of multi-scale models, codec modules and residual block structures    &The model has many parameters and is difficult to train                         \\
		
		SDWNet~\cite{zou2021sdwnet}           & ICCV 2021  & CNN         & A network structure combining extended convolution and wavelet reconstruction is proposed for image deblurring          &Dilated convolution results in a larger perceptual field and reduces the loss of texture information during sampling    &Introduction of wavelet transform increases the complexity of the network       \\
		
		MIMOU-Net~\cite{cho2021rethinking}     & ICCV 2021  & CNN      & Introducing asymmetric feature fusion for gradual recovery of clear images by coder-decoder      & Handling multi-scale blurry with low computational complexity   & Improvement of deblurring accuracy increases computational effort and time consuming     \\
		
		MPRNet~\cite{zamir2021multi}          & CVPR 2021  & CNN         & Multi-stage architecture, injecting supervision at each stage to incrementally improve degraded inputs        & Horizontal connections exist between feature processing blocks to avoid information loss     & High dependence on computing power                                                                                                                               \\
		
		MSSNet~\cite{kim2022mssnet}           & ECCV 2022  & CNN         &A deep learning-based multi-scale stage network from coarse to fine is proposed           &Multi-scale stage configurations and inter-scale information propagation contribute to the network's deblurring performance     &Hyperparameter adjustment is more complex        \\
		
		HINet~\cite{chen2021hinet}          & CVPRW 2022 & CNN         & Propose semi-instantiation to one piece using multi-stage architecture for image recovery task                & HINet architecture generalizes well and has short inference time                     & Requires a large number of image pairs for training                 \\
		
		BANet~\cite{tsai2022banet}            & TIP 2022   & CNN         & Localization of blur direction and size by blurry perception for adaptive blur removal & Short inference time can well support subsequent real-time applications  & Large amounts of diversity data are needed for training to adapt to dynamic scene blurring                                                                       \\
		
		IRNeXt~\cite{cui2023irnext}           & ICML 2023  & CNN         &Integration of multi-stage mechanisms into a u-shaped network to remove different sizes of blur in a coarse-to-fine manner           &The local attention module improves restoration quality by weighting high-frequency information     &The performance improvement of deblurring depends on the number of MSM branches      \\
		
		{MRLPFNet~\cite{dong2023multi} }          & ICCV 2023  & CNN         & A simple and effective multi-scale residual low-pass filtering network           &Effectively model both high-frequency and low-frequency information     &It moderately increases the running time     \\

		SVRNN~\cite{zhang2018dynamic}            & CVPR 2018  & RNN         & End-to-end trainable spatially-variant RNN for dynamic scene deblurring      & The pixel-by-pixel weights of the RNN are learned by the deep CNN, which helps to remove blurring & Need for a larger feeling field, dealing with large areas and spatial variations                                                                                 \\
		
		SRN~\cite{tao2018scale}           & CVPR 2018  & RNN     & Blur removal is achieved using a decoder with skip connections and parameter sharing across three scale         & Sharing network weights at different scales, thus significantly reducing training complexity          & Ignoring the variations between feature scales can easily saturate the network’s performance    \\
		
		MT-RNN~\cite{park2020multi}           & ECCV 2019  & RNN         & Gradual removal of non-uniform blurring over multiple iterations using a self-looping MT structure     & Contextual information is continuous, which facilitates the handling of blurring in dynamic scenes               & Inflexible progressive training may not remove non-uniform blurring very well         \\
		
		{Optical Flow network~\cite{zhang2022deep}}           & TCSVT 2022  & RNN         & Blur removal using optical flow guided spatially variable RNNs     & Optical flow can provide effective information for deblurring dynamic scenes               & Inaccurate optical flow will make it more difficult to recover a clear image        \\

		DeblurGAN~\cite{kupyn2018deblurgan}       & CVPR 2018  & GAN         & Conditional GAN-based network for generating realistic deblurred images                   & Uniform and non-uniform motion blur images with less blurring can be recovered                 & Neglecting the nonlinearity characteristics of dynamic scenes blur, the network sensory field is small and homogeneous             \\
		
		DeblurGAN-V2~\cite{kupyn2019deblurgan}   & ICCV 2019  & GAN         & Use feature pyramid networks and extensive backbone networks for better accuracy          & Improved image restoration efficiency and enhanced real-time performance          & Single network structure, which tends to lead to saturation of the network model \\
		
		DBGAN~\cite{zhang2020deblurring}           & CVPR 2020  & GAN         & Better deblurring by combining learning to generate blurred images and deblurring images    & Synthetic blurred images more accurately mimic real blur for better deblurring            & Network training is complex      \\
		
		{Physics-Based GAN~\cite{pan2021physics} }         & TPAMI 2021  & GAN         & Applied the physical model to generative adversarial networks for image restoration    & The physical model is derived from the image formation process and facilitates image recovery            & Less effective for complex low quality images      \\
		
		CycleGAN~\cite{zhu2017unpaired}        & ICCV 2021  & GAN         & Blind motion deblurring based on multi-adversarial mechanism for iterative generation of high-resolution images          & Add multi-scale edge constraint function to enhance network deblurring and detail preservation capability  & Presence of large color differences, severe artifacts and model redundancy           \\
		
		FCL-GAN~\cite{zhao2022fcl}        & MM 2022   & GAN         & Constraining lightweight network CycleGAN driven tasks by frequency domain contrast losses       & Better deblurring of images in a lightweight and real-time manner & Deblurring effect does not reach the supervised model SOTA level                     \\
		
		Ghost-DeblurGAN~\cite{liu2022application} & IROS 2022  & GAN         & Introducing GhostNet for designing lightweight deblurring networks  & Suitable for real-time image deblurring, small model size, fast inference time           & Deblurring effect fails to reach SOTA level            \\
		
		Uformer~\cite{wang2022uformer}         & CVPR 2021  & Transformer & Building hierarchical codec networks with the Transformer module      & Ability to capture local and global dependencies   & Self-attentive layers can only be applied to a limited depth   \\
		
	{	Restormer~\cite{zamir2022restormer} }     & CVPR 2022  & Transformer &Codecs for learning multi-scale local-global representations of high-resolution images                  & Ability to model global connectivity with high computational efficiency     & Image recovery does not fully utilize spatial information      \\
		
		Stripformer~\cite{tsai2022stripformer}     & ECCV 2022  & Transformer & Re-weighting image features horizontally and vertically to capture blurred features in different directions      & Low memory footprint and computational cost, does not rely on large amounts of training data        & Scaled dot product attention often requires complex matrix multiplication                      \\
		
	{	FFTformer~\cite{kong2023efficient} }     & CVPR 2023  & Transformer & Combining a self-attentive solver with a feed-forward neural network based on discriminative frequency domain to remove blur        & Lower space and time complexity and more effective and efficient in image deblurring         & May not generalize well to different types of images and blurs             \\
		
		BiT~\cite{zhong2023blur}             & CVPR 2023  & Transformer & Introducing Double-Ended Temporal Supervision and Temporal Symmetry Integration Strategies to Generate Effective Time-Varying Motion Rendering Features   & Capable of effectively handling blurring at different scales and fusing information from neighboring frames                 & Limited discrete supervision does not respond well to different realities       \\
		
		Sharpformer~\cite{yan2023sharpformer}    & TIP 2023   & Transformer & Adaptive handling of non-uniform blurring with dynamic convolutional blocks by fully utilizing global-local features       & Able to consistently improve the quality of deblurring due to CNN-based deblurring methods alone         & Higher computational complexity        \\* 
		\bottomrule
	\end{longtable}
\end{landscape}

\section{Datasets and Evaluation Metrics}
\label{sec.dataset}
\subsection{Datasets}
\begin{figure}[t]
	\includegraphics[width=12cm]{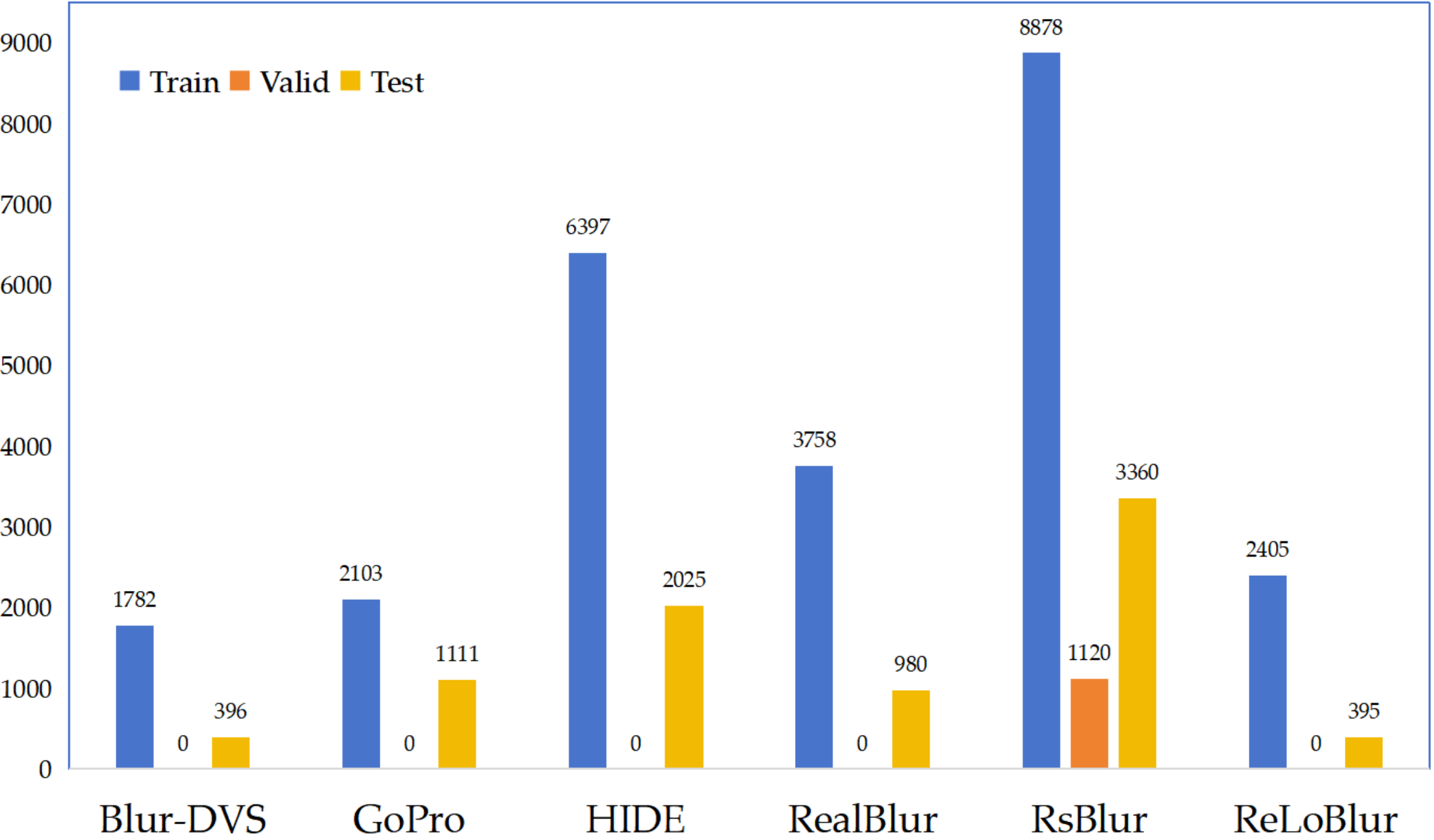}
	\caption{Histogram of the number of samples in the datasets.}
	\label{dataset}
\end{figure}

\begin{figure}[t]
	\centering  % ????
	\includegraphics[width=11cm]{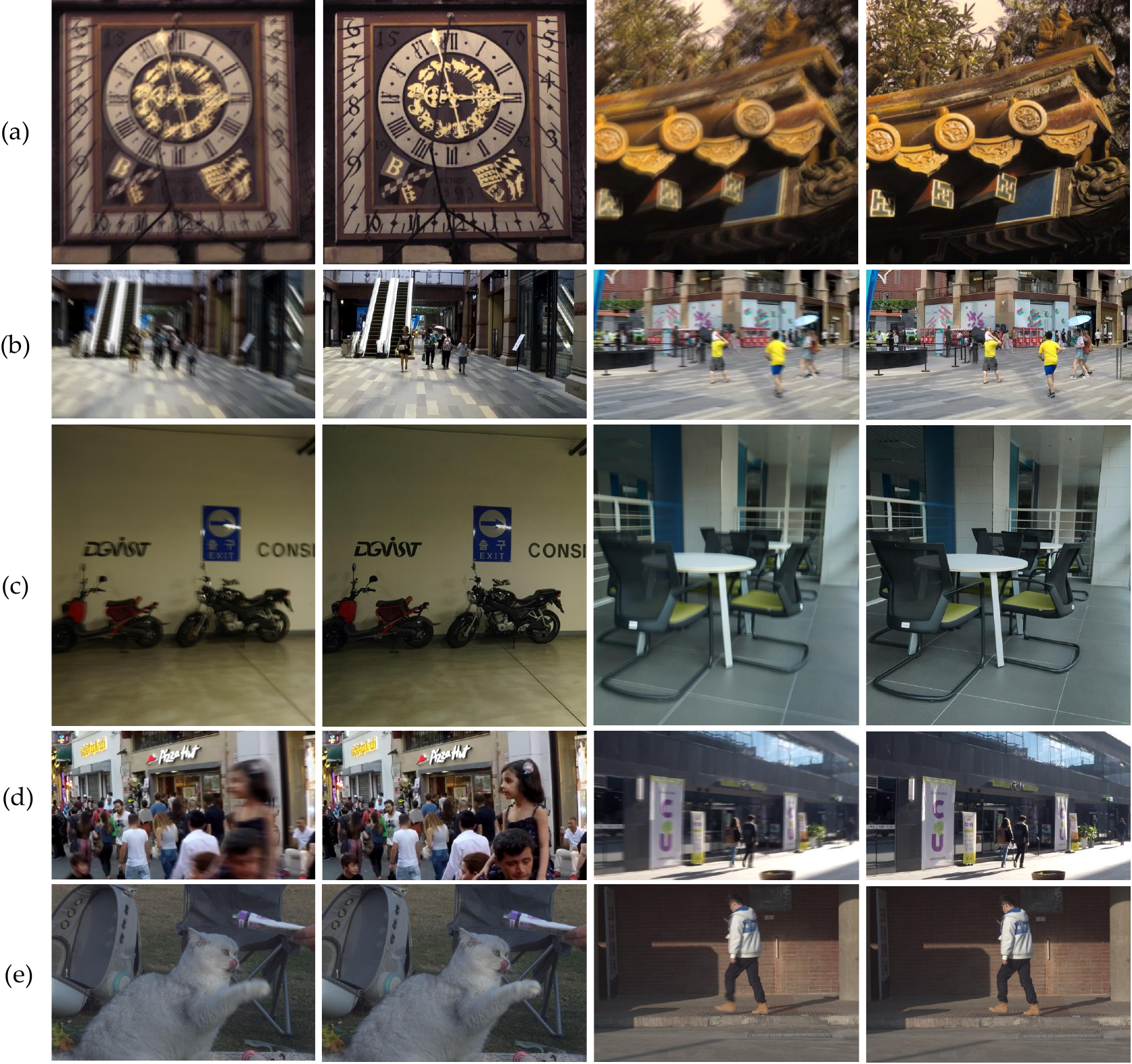}
	\caption{Some example images from (a) K\"{o}hler dataset, (b) HIDE dataset, (c) RealBlur-J dataset, (d) GoPro dataset, (e) ReLoBlur dataset.}
	\label{example_image}
\end{figure}

Most neural network-based image deblurring methods typically require paired images for training. Existing blurry image datasets can be roughly divided into two categories: synthetic datasets and real datasets. {Synthetic datasets usually generate blurred images by frame averaging. While Real datasets capture blurred image using a long exposure time. Synthetic datasets include the K\"{o}hler dataset, Blur-DVS dataset, GoPro dataset, and HIDE dataset, among others. } Real image datasets comprise the RealBlur dataset, RsBlur dataset, ReLoBlur dataset, and so on. As shown in \figurename~\ref{example_image}, these are example images from some commonly used datasets. \tablename~\ref{tab:datasets} summarizes and outlines the basic information of select datasets. We additionally provide bar graphs to compare the size of the datasets, as shown in \figurename~\ref{dataset}.

\textbf{K\"{o}hler et al. Dataset}: To simulate real camera shake, K\"{o}hler et al.~\cite{kohler2012recording} used a Stewart platform with six degrees of freedom to replay the camera's motion trajectory. During the replay of the motion trajectory, a series of clear images and long-exposure blurry images were captured. The dataset comprises 4 clear images and 48 corresponding blurry images for 12 motion trajectories.

%\begin{adjustwidth}{-1in}{-1in} % 负值表示表格周围的页边距
	\begin{table}[b]
%		\tiny 
		\centering
		\caption{Representative benchmark datasets for evaluating single image deblurring algorithms.}
		\label{tab:datasets}
		\setlength{\tabcolsep}{0.1mm}{%
		\begin{tabular}{clllc}
			\toprule
			\textbf{Real/Syn.} & \textbf{Dataset} & \textbf{Generative approach} & \textbf{Sizes} & \textbf{Train/Valid/Test} \\ \midrule
			\multirow{4}{*}{Synthetic} & K\"{o}hler at al.~\cite{kohler2012recording} & motion trajectory captured by 6D camera & 4 sharp+48 blur &	Not divided \\
			& GoPro~\cite{nah2017deep} & averaging over consecutive clear frames  & 3,214 pairs & 2,103 / 0 / 1,111 \\
			& HIDE~\cite{shen2019human} & averaging over consecutive clear frames & 8,422 pairs & 6,397 / 0 / 2,025 \\
			& Blur-DVS~\cite{jiang2020learning} & averaging over consecutive clear frames  & 2178 pairs & 1782/ 0 / 396 \\  
			\midrule
			\multirow{3}{*}{Real} 
			& RealBlur~\cite{rim2020real} & \begin{tabular}[l]{@{}c@{}}blurred images: low shutter speed cameras \\ clear images: high shutter speed cameras\end{tabular} & 4,738 pairs & 3,758 / 0 /980 \\
			& RsBlur~\cite{rim2022realistic}   & \begin{tabular}[c]{@{}l@{}}blurred images: long exposure camera\\ clear images:  short exposure camera\end{tabular} & 13,358 pairs & 8,878 / 1,120 / 3,360 \\
			& ReLoBlur~\cite{li2023real} & \begin{tabular}[c]{@{}l@{}}blurred images: long exposure camera\\ clear images: short exposure time camera\end{tabular} & 2,405 pairs & 2,010 / 0 /395 \\ 
			\bottomrule
		\end{tabular}
	}
	\end{table}
%\end{adjustwidth}

\textbf{Blur-DVS Dataset}: Created by Jiang et al.~\cite{jiang2020learning} using the DAVIS240C camera. The creation process involved capturing an image sequence with a slow-motion camera and then synthesizing 2,178 pairs of blurred and clear images by averaging adjacent 7 frames. The dataset consists of 1,782 pairs for training and 396 pairs for testing. Additionally, the dataset provides 740 real blurry images without paired clear images.

\textbf{GoPro Dataset}: Developed by Nah et al.~\cite{nah2017deep}, this dataset is one of the most commonly used deep learning methods. The dataset's motion-blurred images are generated by integrating multiple instantaneous clear images within a time interval. Using a GoPro Hero4 Black camera capturing clear images at 240fps, they averaged over varying time windows to create blurred images, with the clear image at the window's center serving as the ground truth. It includes 3,214 pairs of clear and blurred images, divided into 2,103 pairs for training and 1,111 pairs for testing.

\textbf{HIDE Dataset}: Introduced by Shen et al.~\cite{shen2019human}, this dataset focuses on motion blur in pedestrian and street scenes, including camera shake and object movement. The dataset was created by capturing videos at 240 fps using a GoPro Hero camera and then averaging frames from these high-frame-rate videos. They averaged 11 consecutive frames to generate blurred images and kept the center frame as the clear image. It comprises 8,422 clear-blurred image pairs, divided into 6,397 pairs for training and 2,025 pairs for testing.

\textbf{RealBlur Dataset}: Rim et al.~\cite{rim2020real} created the RealBlur dataset to train and test deep deblurring methods on real blurry images. The dataset consists of two subsets sharing the same image content. One subset, RealBlur-J, is generated from the camera's raw images, while the other, RealBlur-R, is generated from JPEG images processed by the camera's ISP. It contains a total of 9,476 image pairs, with each subset offering 4,556 pairs of blurry and ground truth clear images for 232 low-light static scenes. The blurry images in the dataset are affected by camera shake and were captured in low-light scenarios such as night streets and indoor rooms, covering common motion blur scenes.

\textbf{RsBlur Dataset}: Developed by Jaesung Rim et al.~\cite{rim2022realistic}, the RsBlur dataset provides real blurry images of various outdoor scenes, each paired with a series of 9 clear images. It comprises 13,358 real blurry images from 697 scenes. For analysis, the dataset is divided into training, validation, and testing sets, containing 8,878, 1,120, and 3,360 blurry images, respectively, with 465, 58, and 174 scenes in each set.

\textbf{ReLoBlur Dataset}: Designed by Li et al.~\cite{li2023real}, ReLoBlur uses a simultaneous multi-band shutter device to capture local motion-blurred images along with their corresponding clear images, followed by post-processing for correction. Unlike synthetic blur datasets, ReLoBlur authentically showcases the mixing effect between locally moving objects and backgrounds, including overexposed, dimly lit real-life scenes, and complex blurred edges. It contains common real-life scenes such as pedestrians, vehicles, family members, pets, balls, plants, and furniture, among others. The dataset publicly provides 2,010 pairs of training set images and 395 pairs for testing. Additionally, the authors annotated the areas of local motion blur.

\subsection{Evaluation Metrics}
Image quality assessment (IQA) is an important method to evaluate the effectiveness of the model. Current image quality evaluation methods can be divided into subjective and objective evaluation indexes.
\subsubsection{Subjective Evaluation}
Subjective evaluation relies on human judgment and is often the most intuitive method for assessing image quality. One representative measure is the Mean Opinion Score (MOS)~\cite{hossfeld2016qoe}. Subjective evaluation scales primarily include absolute and relative scales. 
As shown in \tablename~\ref{tab:subjective}, individuals rate image quality within the range of 1-5. However, when different deblurring methods produce similar results, subjective evaluation often struggles to differentiate subtle differences. 
This susceptibility to individual biases and external environmental factors during assessment can lead to varying subjective evaluation results among different observers. Therefore, subjective evaluation is generally used as a supplementary assessment method in image quality evaluation and cannot entirely determine the final image quality assessment outcome. 
As a result, for image deblurring, most existing methods rely on objective evaluation metrics for assessment.

\begin{table}[t]
	\caption{Subjective evaluation scale.}
	\label{tab:subjective}
%\tiny 
		%		\begin{tabular}{\fulllength}{m{1cm}<{\centering}CCm{3.5cm}<{\centering}CC}
		\setlength{\tabcolsep}{0.005mm}{%
			\begin{tabular}{cccc}
			%\begin{tabular}{clll}
				\toprule
				\textbf{Score $ \uparrow $}	& \textbf{Quality scale}	& \textbf{Relative quality}     & \textbf{Absolute quality} \\
				\midrule
				\multirow[m]{1}{*}{5}	& There is no sign of the image quality deteriorating.			& The best.			& Beyond compare\\
				
				\multirow[m]{1}{*}{4}	& The image quality is poor, but it is still viewable.			& Better than the average level.			& Nice\\
				
				\multirow[m]{1}{*}{3}	& The image quality is poor, slightly hampers viewing.			& The average level.			& Commonly\\
				
				\multirow[m]{1}{*}{2}	& Hindrance to watching.										& Below average in favorability.			& Bad\\
				
				\multirow[m]{1}{*}{1}   & Very serious obstruction of viewing.							& The worst  & Very bad \\
				\bottomrule
			%\end{tabular}
	\end{tabular}
}
\end{table}

\subsubsection{Objective Evaluation}
Objective assessment primarily involves quantitatively analyzing and evaluating deblurred images through the construction of mathematical models~\cite{su2022survey,li2022survey,wang2004image}. 
Objective assessment enables the unified measurement of various deblurring methods through precise data calculations, offering advantages of fairness, objectivity, and efficiency. 
Among these evaluation methods, Mean Squared Error (MSE), Peak Signal-to-Noise Ratio (PSNR), and Structural Similarity Index (SSIM) stand as the most commonly employed metrics in image restoration tasks.

\textbf{Mean Square Error (MSE).}
\textsc{MSE} is the average of the squared absolute errors between the restored image and the ground-truth sharp image. The closer the MSE is to zero, the smaller the difference between the restored image and the ground truth image, indicating a more realistic restoration~\cite{li2022survey}.
	\begin{equation}
			\operatorname{MSE}(\mathbf{I}, \mathbf{S}) = \frac{1}{K} \sum_{k=0}^{K-1}\left(\mathbf{I}_k-\mathbf{S}_k\right)^2,
		\end{equation}
		where $\mathbf{I}$ and $ \mathbf{S} $ represent the reconstructed image and the real sharp image, respectively.
		$ K $ denotes all pixels in the image.
		
		\textbf{Peak Signal Noise Ratio (PSNR).}
		PSNR~\cite{hore2010image} is a commonly used metric to assess the effectiveness of image deblurring. 
		It measures the accuracy of image deblurring by comparing the MSE between the deblurred and sharp images. However, PSNR only focuses on differences at the pixel level. In image deblurring assessment, it's typical to combine other metrics to comprehensively evaluate image quality. PSNR is defined as:
		
		\begin{equation}
			\operatorname{PSNR}(\mathbf{I}, \mathbf{S}) = 10 \log _{10}\left(\frac{\left(2^n-1\right)^2}{\operatorname{MSE(\mathbf{I}, \mathbf{S})}}\right),
		\end{equation}
		where $n$ is the number of digits where the image is stored. The larger the PSNR value is, the closer the recovered image is to the original image, the better the image recovery is.
		
\begin{table}[b]
%			\tiny 
			\centering
			\caption{
				%			Performance evaluation of representative methods for image deblurring on four popular image deblurring	datasets.
				{Performance comparison of state-of-the-art methods on four commonly used image deblurring datasets: GoPro, HIDE, RealBlur-J, and RealBlur-R.
				In {\color{red}\textbf{red}} are depicted the overall best results among all methods, while in \textbf{bold} are showcased the best results within each respective category.}
			}
			\label{tab:sota}
			%	\resizebox{\textwidth}{!}{%
				\setlength{\tabcolsep}{0.3mm}{%
					\begin{tabular}{llcccccccc}
						\toprule
						\multirow{2}{*}{\textbf{Methods}} & \multirow{2}{*}{\textbf{Category}} & \multicolumn{2}{c}{\textbf{GoPro}~\cite{nah2017deep}} & \multicolumn{2}{c}{\textbf{HIDE}~\cite{shen2019human}} & \multicolumn{2}{c}{\textbf{RealBlur-J}~\cite{rim2020real}} & \multicolumn{2}{c}{\textbf{Realblur-R}~\cite{rim2020real}} \\ \cmidrule(l){3-4}  \cmidrule(l){5-6}  \cmidrule(l){7-8}  \cmidrule(l){9-10}  
						&   & \textbf{PSNR}        & \textbf{SSIM}   & \textbf{PSNR}    & \textbf{SSIM}   & \textbf{PSNR}       & \textbf{SSIM}     & \textbf{PSNR}    & \textbf{SSIM}     \\ \midrule
						DeepDeblur~\cite{nah2017deep}       & CNN                       & 29.08       & 0.914       & 25.73       & 0.874      & 27.87          & 0.834         & 32.51          & 0.841         \\
						DMPHN~\cite{zhang2019deep}           & CNN                       & 31.20       & 0.940       & 29.09       & 0.924      & 28.42          & 0.860         & 35.70          & 0.948         \\
						SDWNet~\cite{zou2021sdwnet}           & CNN                       & 31.26       & 0.966       & 28.99       &\color{red}\textbf{0.957}      & 28.61          & 0.867         & 35.85          & 0.948         \\
						MSCAN~\cite{wan2020deep}           & CNN                     & 31.24      & 0.945    & 29.63    & 0.920 & -     & -    & -     & -      \\			
						PSS-NSC~\cite{gao2019dynamic}           & CNN          & 31.58     & 0.948     & -       & -      & -          & -         & -          &-       \\
						MIMOU-Net+~\cite{cho2021rethinking}      & CNN                       & 32.45       & 0.957       & 29.99       & 0.930      & 27.63          & 0.837         & 35.54          & 0.947         \\
						MPRNet~\cite{zamir2021multi}           & CNN                       & 32.66       & 0.959       & 30.96       & 0.939      & 28.70          & 0.873         & \textbf{35.99}          & 0.952         \\
						HINet~\cite{chen2021hinet}            & CNN                       & 32.77       & 0.959       & -           & -          & -              & -             & -              & -             \\
						BANet~\cite{tsai2022banet}           & CNN                       & 32.54       & 0.957       & 30.16       & 0.930      & -         & -          & -         & -         \\ 
						SAPHNet~\cite{suin2020spatially}           & CNN                       & 32.02       & 0.953       & 29.98       & 0.930      & -         & -         & -         & -         \\
						MSSNet~\cite{kim2022mssnet}           & CNN    & 33.01    & 0.961   & 30.79    & 0.939    & \textbf{28.79}   & \textbf{0.879} & 35.93         & 0.953         \\
						{MSFS-Net~\cite{zhang2023multi}}  & CNN  & 32.73 & 0.959 & 31.05       & 0.941      &  28.97         & 0.908          &  36.02          &  0.959          \\  
					
						LaKDNet~\cite{ruan2023revisiting}  & CNN  & 33.35 & 0.964 &31.21      & 0.943      & 28.78        & 0.878          & 35.91         & \textbf{0.954}         \\  
						{MRLPFNet~\cite{dong2023multi} } & CNN  & \textbf{33.50} & \textbf{0.965} & \color{red}{\textbf{31.63}  }     & 0.946      &  -         & -          &  -          &  -          \\   \midrule
											
						SRN~\cite{tao2018scale}               & RNN                   & 30.26       & 0.934       & 28.36       & 0.915      & 28.56          & 0.867         & 35.66          & 0.947         \\
						MT-RNN~\cite{park2020multi}          & RNN                       & 31.15       & 0.945       & 29.15       & 0.918      & 28.44          & 0.862         & 35.79          & 0.951         \\ 
						ESTRNN~\cite{zhong2023real}          & RNN        & 31.07       & 0.902       & -       & -      & -     & -       & -    & -        \\
						IFI-RNN~\cite{nah2019recurrent}          & RNN        & 29.97       & 0.895       & -       & -      & -     & -       & -    & -        \\
						RNN-MBP~\cite{zhu2022deep}          & RNN  & \textbf{33.32}    & \textbf{0.963}       & -       & -      & -     & -       & -    & -        \\ \midrule
						DeblurGAN~\cite{kupyn2018deblurgan}       & GAN                       & 28.70       & 0.858       & 24.51       & 0.871      & 27.97          & 0.834         & 33.79          & 0.903         \\
						DeblurGAN-V2~\cite{kupyn2019deblurgan}     & GAN    & 29.55       & 0.934       & 26.61       & 0.875      & \textbf{28.70} & \textbf{0.866}    & \textbf{35.26} & \textbf{0.944}         \\
						DBGAN~\cite{zhang2020deblurring}             & GAN  & \textbf{31.10}  & \textbf{0.942}  & \textbf{28.94}   & \textbf{0.915}      & 24.93          & 0.745         & 33.78          & 0.909         \\
						CycleGAN~\cite{zhu2017unpaired}          & GAN                       & 22.54       & 0.720       & 21.81       & 0.690      & 19.79          & 0.633         & 12.38          & 0.242         \\
						FCL-GAN~\cite{zhao2022fcl}         & GAN                       & 24.84       & 0.771       & 23.43       & 0.732      & 25.35          & 0.736         & 28.37          & 0.663         \\
						Ghost-GAN~\cite{liu2022application} & GAN                       & 28.75       & 0.919       & -           & -          & -              & -             & -              & -             \\ \midrule
						CTMS~\cite{zhao2022rethinking}          & Transformer               & 32.73       & 0.959       & 31.05       &0.940      & 27.18          & 0.883         & -        & -       \\	
						Uformer~\cite{wang2022uformer}          & Transformer               & 32.97       & 0.967       & 30.83       & \textbf{0.952 }     &\color{red}\textbf{ 29.06 }         & \color{red}\textbf{0.884 }        & \color{red}\textbf{36.22  }        & \color{red}\textbf{0.957  }       \\
						{Restormer~\cite{zamir2022restormer} }        & Transformer               & 32.92       & 0.961       &31.22      & 0.942      & 28.96          & 0.879         & 36.19          & 0.957         \\
						Stripformer~\cite{tsai2022stripformer}     & Transformer               & 33.08       & 0.962       & 31.03       & 0.940      & -         & -         & -          &-      \\
						Stoformer~\cite{xiao2022stochastic}      & Transformer               & 33.24       & 0.964       & 30.99       & 0.941     & -         &-        & -        & -         \\
						{FFTformer~\cite{kong2023efficient} }        & Transformer               & \color{red} \textbf{34.21}       & \color{red}\textbf{0.969}       & \textbf{31.62}       & 0.945      & -         &-        & -        & -         \\
						Sharpformer~\cite{yan2023sharpformer}     & Transformer               & 33.10       & 0.963       & -           & -          & -              & -             & -              & -             \\ 
						\bottomrule
					\end{tabular}
				}
				
			\end{table}

		\textbf{Structural Similarity Index Measure(SSIM).}
		SSIM~\cite{wang2004image} is one of the commonly used metrics in image quality assessment. In the task of image deblurring, it evaluates the similarity between the deblurred image and the ground-truth sharp image by comparing their structural information. SSIM primarily assesses image similarity based on three characteristics: luminance, contrast, and structure. It doesn't just focus on pixel differences but also considers features like image structure and texture. Compared to PSNR, SSIM aligns better with human visual perception. The calculation formula is as follows:
		
		\begin{equation}
			\operatorname{SSIM}(\mathbf{I}, \mathbf{S}) = [l(\mathbf{I}, \mathbf{S})]^\alpha \cdot [c(\mathbf{I}, \mathbf{S})]^\beta \cdot [s(\mathbf{I}, \mathbf{S})]^\gamma,
			\label{Eq.ssim}
		\end{equation}
		where $l(\mathbf{I}, \mathbf{S})$ represents the brightness comparison of the image,
		$c(\mathbf{I}, \mathbf{S})$ represents the contrast comparison of the image,
		$s(\mathbf{I}, \mathbf{S})$ represents the structural comparison of the image.
		In engineering, often make $\alpha=\beta=\gamma=1$.
		In Eq.~\eqref{Eq.ssim},
		\begin{equation}
			l(\mathbf{I}, \mathbf{S})= \frac{2 \mu_I \mu_S+C_1}{\mu_I^2+\mu_S^2 + C_1},~~~
			c(\mathbf{I}, \mathbf{S}) = \frac{2 \sigma_I \sigma_S+C_2}{\sigma_I^2+\sigma_S^2 + C_2},~~~
			s(\mathbf{I}, \mathbf{S}) = \frac{\sigma_{I S}+C_3}{\sigma_I \sigma_S + C_3},
			\label{Eq.lcs}
		\end{equation}
		where $\mu_I$, $\sigma_I$ and $\mu_S$, $\sigma_S$ represent the mean, standard deviation of the images $ \mathbf{I}$ and $\mathbf{S} $, respectively.
		$\sigma_{IS} = \frac{1}{K} \sum_{k=0}^{K-1}\left(\mathbf{I}_k - \mu_I\right)\left(\mathbf{S}_k - \mu_S\right) $ indicates the covariance of the reconstructed image $ \mathbf{I} $ and the real sharp image $ \mathbf{S} $.
		$C$ is a constant to avoid the denominator 0 causing system errors.
		Both $C_1$, $C_2$ and $C_3$ are constants, with values ranging from $ [-1, 1] $. A higher SSIM value indicates a greater similarity between the restored image and the original clear image, signifying a better restoration outcome.

	\begin{figure}[b]
		
		\centering  % ????
		\includegraphics[width=13cm]{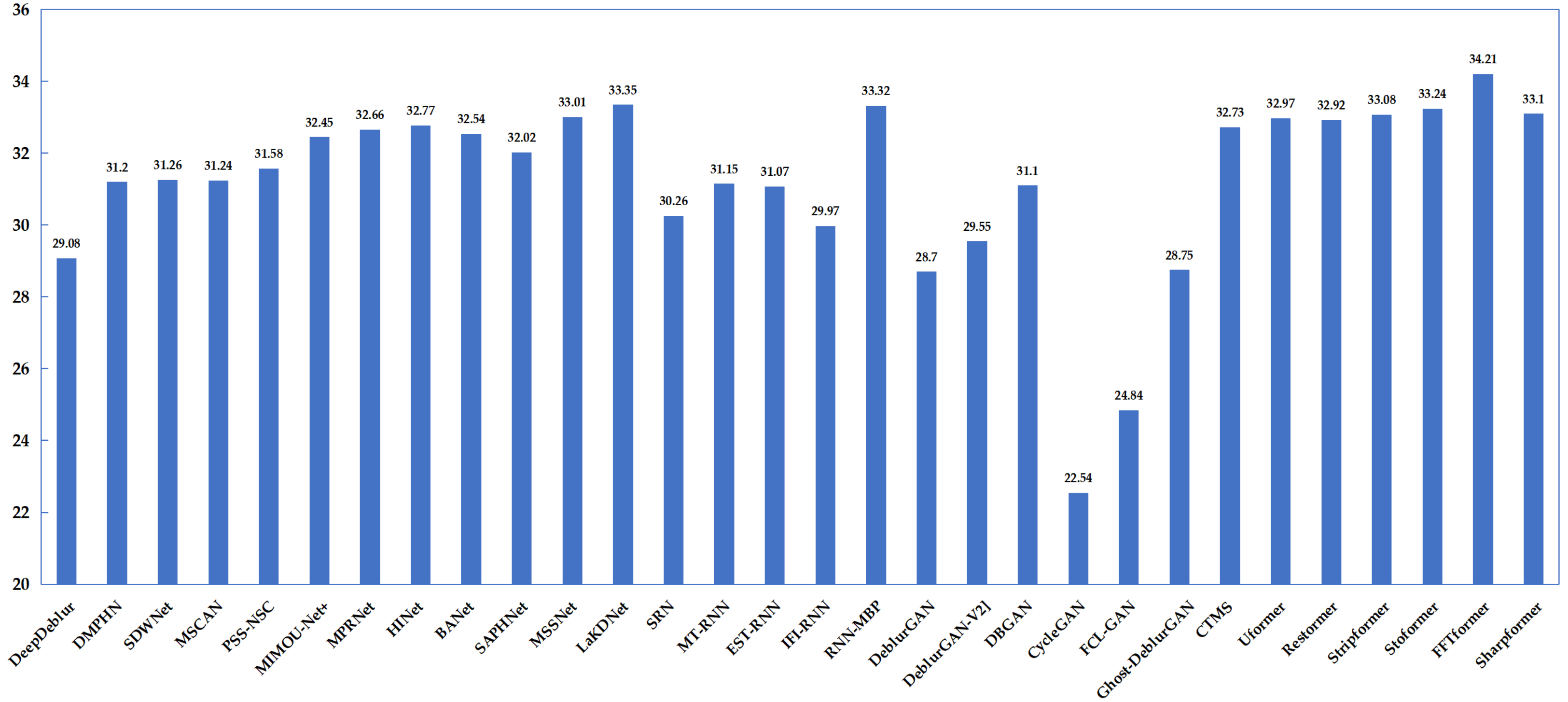}
		
		\caption{
			%		Comparison among image deblurring methods in terms of PSNR on the GoPro dataset.
			Performance comparison (PSNR, \%) of SOTA methods on the GoPro dataset.
		}	\label{psnr}
	\end{figure}

\section{Comparison with State-of-the-Art Methods}
\label{sec.methods}
\subsection{Quantitative Comparison.}
Quantitative comparisons were conducted among 30 blind motion deblurring methods across four datasets, encompassing 12 CNN-based, 5 RNN-based, 6 GAN-based, and 7 Transformer-based approaches. 
From \tablename~\ref{tab:sota}, it is evident that Transformer-based methods exhibit competitive performance across the majority of datasets. This can be attributed to the self-attention mechanism in Transformers, which adeptly captures remote dependencies between local and global image sequences, enhancing the model's ability to recover details from blurred images.
As depicted in \tablename~\ref{tab:sota}, minimal performance disparities were observed between CNN and RNN methods on the GoPro dataset, both achieving over 33\% PSNR and 0.96 SSIM. 
This similarity can be attributed partly to their analogous structures and the relatively simple motion blur in the synthetic nature of the GoPro dataset.
\figurename~\ref{psnr} illustrates that FFTformer achieved the highest PSNR score at 34.21\%. 
Conversely, as indicated in \figurename~\ref{psnr}, the performance of GAN-based methods was inferior to other types due primarily to challenges related to training difficulties and convergence issues within GAN models, resulting in suboptimal deblurring restoration effects.

\begin{table}[t] 
	\caption{Speed and Parameters of representative SOTA methods.
		%The numbers are quoted from~\cite{kong2023efficient}.
	}
	\label{tab:param}
	\newcolumntype{C}{>{\centering\arraybackslash}X}
	\begin{tabularx}{\textwidth}{lCCCC}
		\toprule
		\textbf{Method}	& \textbf{Category} & \textbf{Param. (M)}	& \textbf{Time (s)} & \textbf{PSNR (\%)}\\
		\midrule
		DMPHN~\cite{zhang2019deep}					& CNN			&21.7				& 0.21		& 31.20 \\
		MIMOU-Net+~\cite{cho2021rethinking}			& CNN			&16.1				& \textbf{0.02}				& 32.45 \\
		MPRNet~\cite{zamir2021multi}				& CNN			&23.0				&0.09				& 32.66 \\
		SRN~\cite{tao2018scale}						& RNN			& \textbf{6.8}		&0.07				& 30.26 \\
		DeblurGAN-V2~\cite{kupyn2019deblurgan}   	& GAN			&60.9				&0.04  				& 29.55 \\
		Restormer~\cite{zamir2022restormer} 		& Transformer	&26.1				&0.08				& 32.92 \\
		Uformer~\cite{wang2022uformer} 				& Transformer	&50.9				&0.07				& 32.97 \\
		Stripformer~\cite{tsai2022stripformer}		& Transformer	&19.7				&0.04				& 33.08 \\
		FFTformer~\cite{kong2023efficient}			& Transformer	&16.6				&0.13				& \textbf{34.21} \\
		\bottomrule
	\end{tabularx}
\end{table}

In addition, \tablename~\ref{tab:param} presents a comparative analysis of various methods, balancing parameter count and operational speed. 
It is evident from \tablename~\ref{tab:param} that RNN-based methods exhibit the smallest parameter count, CNN-based methods demonstrate the fastest operational speed, while Transformer-based methods yield the best performance. 
This insight further guides us toward integrating the strengths of different methodological types in subsequent research endeavors, aiming to achieve high-performance, real-time blind motion deblurring methods.

\begin{figure}[h]
	\centering  % ????
	\includegraphics[width=13cm]{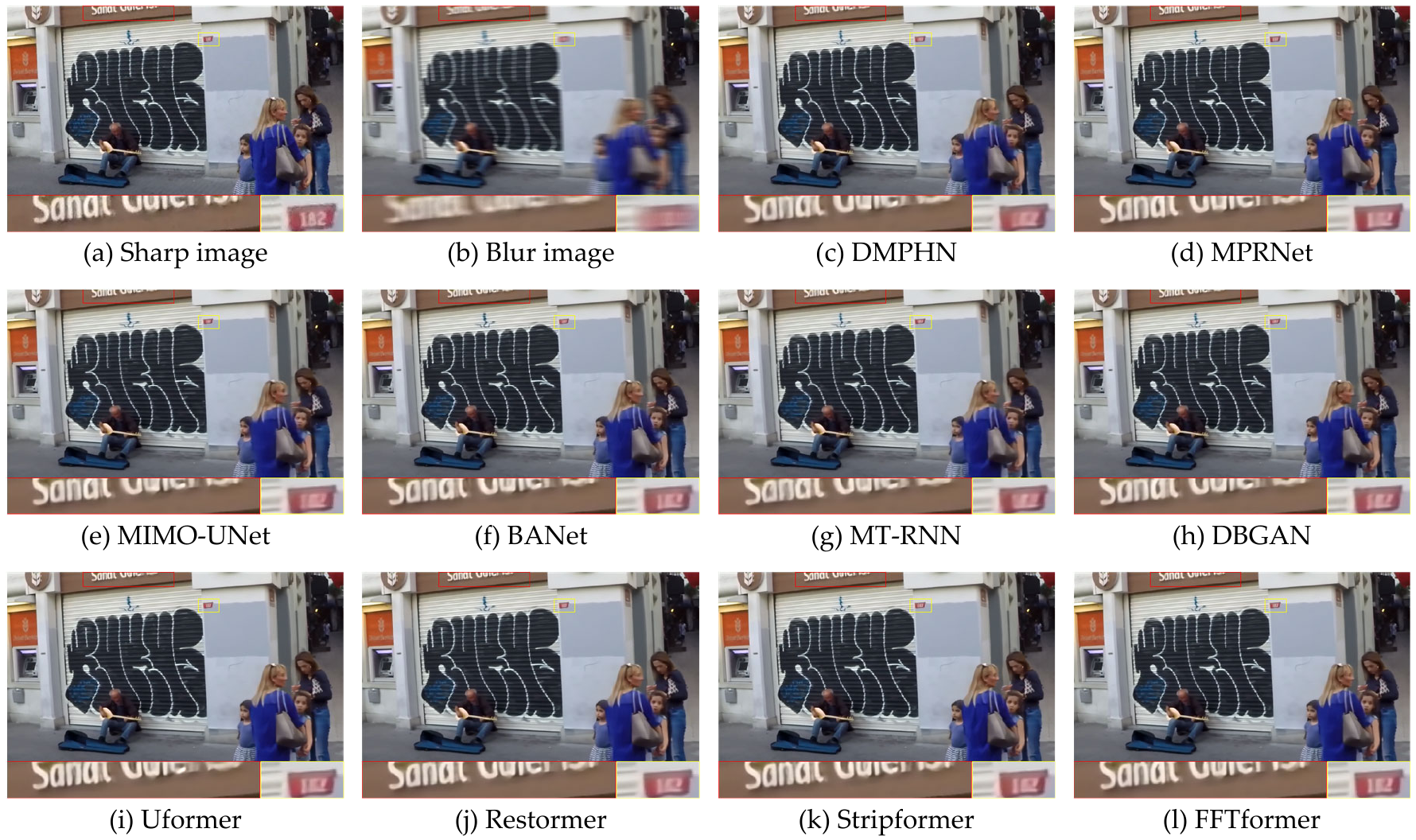}
	\caption{Comparison of background deblurring results on the GoPro dataset.}
	\label{performance_background}
\end{figure}

\begin{figure}[t]
	\centering  % ????
	\includegraphics[width=13cm]{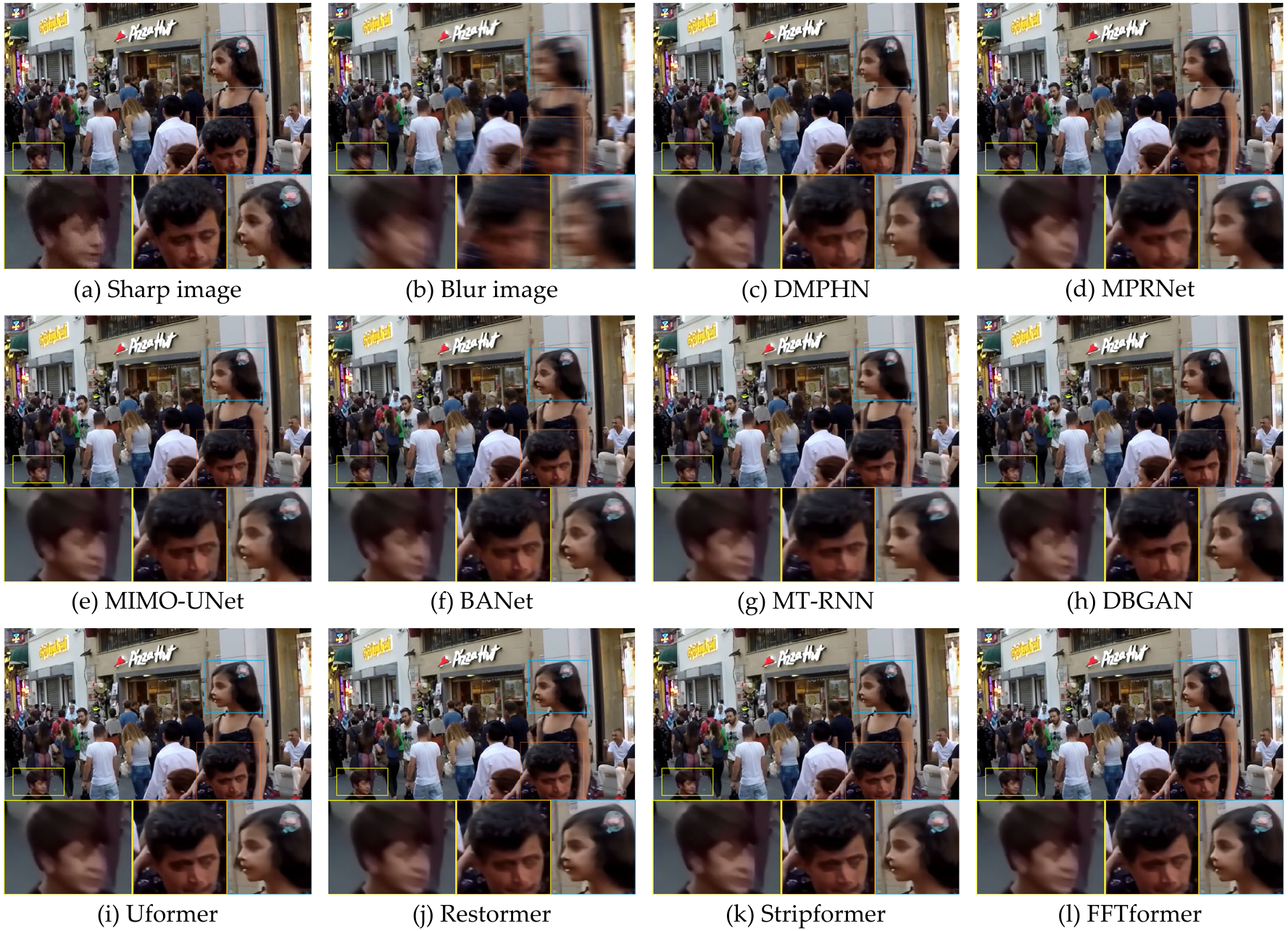}
	\caption{Comparison of human deblurring results on the GoPro dataset.}
	\label{performance_human}
\end{figure}

\subsection{Qualitative Comparison.}
As shown in \tablename~\ref{tab:advantages}, this review summarizes over 30 state-of-the-art blind motion deblurring methods based on deep learning. It includes publication details, summaries, advantages, and limitations. 
Among these, 22 were published in top-tier computer vision conferences such as CVPR, and ICCV, while 8 were featured in leading image processing journals like TIP, and TCSVT, among others.

To visually compare the deblurring performance of different algorithms, we conducted a qualitative assessment of selected algorithmic performance on the GoPro dataset. 
\figurename~\ref{performance_background} and \figurename~\ref{performance_human} illustrate the deblurring effects of various methods on backgrounds and human subjects, respectively.
DBGAN and Uformer depict the deblurring results published by the original authors, while the remaining deblurred results were reproduced by us. The experiments were conducted on a Windows 10 system using Python 3.8 and Torch 2.0.0.
As depicted in \figurename~\ref{performance_background} and \figurename~\ref{performance_human}, the deep learning-based blind image deblurring methods tested exhibit effective image restoration. While some differences persist compared to the original clear images, visually, they do not significantly impact the image information. Combining the insights from Table 4 and Figure 13, it becomes apparent that Transformer-based models not only outperform other methods in evaluation metrics such as PSNR and SSIM but also demonstrate superior restoration of image details throughout the image recovery process.

%%%%%%%%%%%%%%%%%%%%%%%%%%%%%%%%%%%%%%%%%%
\section{Current Challenges and Future Prospects}
\label{sec.challenge}
%\subsection{Challenges of Blind Motion Deblurring and Major Concerns}
The paper provides an overview of image deblurring algorithms based on deep learning, encompassing methods such as CNN-based, RNN-based, GAN-based, and Transformer-based image deblurring techniques. These methods are analyzed, compared, and summarized, and their performance is contrasted.

Despite the progress made in image deblurring algorithms concerning restoration effects, challenges persist due to the complexity of real-world blur scenarios. The challenges of blind motion deblurring and future prospects are as follows:

(1) Balancing Model Performance and Light-weight Models: Increasing model depth and parameters can enhance fitting data and improve image restoration quality. However, this can lead to larger network sizes, longer inference times, and higher computational complexity. Finding a balance between model performance and model light-weighting for practical applications is an essential area for exploration.

(2) Deblurring Methods for High-Level Visual Tasks:
Blurred images hinder several advanced computer vision tasks, yet many existing deblurring methods aim solely at recovering high-quality images without considering aiding higher-level tasks such as object detection~\cite{li2023detection,wu2023deep,zhao2023attacking} or image segmentation~\cite{pan2016soft,luo2021blind}.
Research focusing on effectively extracting relevant features for high-level visual problems to enhance image deblurring solutions is a promising direction.

(3) Few-shot Learning: Traditional deep learning models often require extensive labeled data for training, which might be challenging to obtain, especially high-quality labeled clear images. Few-shot learning~\cite{zhang2019few,chi2021test,chi2021test} help models rapidly learn and adapt to new tasks from a limited amount of labeled data, reducing dependence on extensive labeled datasets. This approach addresses challenges related to acquiring and labeling data, potentially widening the application of image deblurring techniques across real-world scenarios.

(4) Applications of Diffusion: The Diffusion model is a generative model that has garnered significant attention in recent years, demonstrating immense potential in the realm of image editing and enhancement. In the field of image deblurring, the Diffusion model has also found applications~\cite{ren2023multiscale,whang2022deblurring}. Its flexibility allows it to handle various types of noise and transformations, suggesting its capability to potentially address complex image blurring. In the future, with a deeper understanding of the Diffusion model and the development of related techniques, it is poised to become a significant tool in the domain of image deblurring, promising more satisfying results and improvements.

\section{Conclusions}
\label{sec.conclusion}
The current research and developments indicate significant progress in enhancing image quality and clarity through deep learning technologies. This paper reviews recent advancements in deep learning-based image deblurring algorithms, providing a comprehensive classification and analysis. It elaborates on existing image deblurring algorithms from four perspectives: CNN, RNN, GAN, and Transformer. Additionally, it conducts comparative analyses of representative algorithms, summarizing their strengths and limitations. Finally, the paper outlines future research directions in image deblurring algorithms. However, it is acknowledged that addressing more complex blurs in real-world scenarios remains a challenge. The quality of data and the accuracy of labels are crucial for training deep learning models, necessitating large-scale, high-quality datasets to support model training and optimization. In the future, there is an anticipation to further optimize deep learning models to enhance their speed and efficiency, expanding the application of these techniques in fields such as video processing, surveillance, and autonomous driving.All collected models, benchmark datasets, source code links, and codes for evaluation have been made publicly available at \url{https://github.com/VisionVerse/Blind-Motion-Deblurring-Survey}.

\section*{Declarations}

\begin{itemize}
\item Funding:This research received no external funding.
\item Author contribution:	Methodology, Y.X.; 
Supervision, Z.L. and Y.X.; 
Writing—Original Draft, Y.X., H.Z. and C.L.; 
Writing—Review and Editing, Y.X., H.Z. and F.S.;
Visualization, Y.X., F.S., C.L. and H.Z.;
All authors have read and agreed to the published version of the manuscript.
\end{itemize}

\bibliography{sn-bibliography}% common bib file
%% if required, the content of .bbl file can be included here once bbl is generated
%%\input sn-article.bbl
\balance

\end{document}